# Vision Transformers, a new approach for high-resolution and large-scale mapping of canopy heights

Ibrahim Fayad[a,b*], Philippe Ciais[a], Martin Schwartz[a], Jean-Pierre Wigneron[c], Nicolas Baghdadi[d], Aurélien de Truchis[b], Alexandre d'Aspremont[e,b], Frederic Frappart[c], Sassan Saatchi[f], Agnes Pellissier-Tanon[a], Hassan Bazzi[g]

(a) Laboratoire des Sciences du Climat et de l'Environnement, LSCE/IPSL, CEA-CNRS-SCE32UVSQ, Université Paris Saclay, 91191 Gif-sur-Yvette, France.
(b) Kayrros SAS, Paris, 75009, France
(c) ISPA, UMR 1391, INRA Nouvelle-Aquitaine, Bordeaux Villenave d'Ornon, France
(d) INRAE, UMR TETIS, Université de Montpellier, AgroParisTech, CEDEX 5, 34093 Montpellier, France
(e) Department of Computer Science, École Normale Supérieure, Paris 75230, France
(f) Jet Propulsion Laboratory (JPL), California Institute of Technology, Pasadena, CA, 91125, USA
(g) Université Paris-Saclay, AgroParisTech, INRAE, UMR 518 MIA Paris-Saclay, 91120 Palaiseau, France

**Abstract**

Accurate and timely monitoring of forest canopy heights is critical for assessing forest dynamics, biodiversity, carbon sequestration as well as forest degradation and deforestation. Recent advances in deep learning techniques, coupled with the vast amount of spaceborne remote sensing data offer an unprecedented opportunity to map canopy height at high spatial and temporal resolutions. Current techniques for wall-to-wall canopy height mapping correlate remotely sensed information from optical and radar sensors in the 2D space to the vertical structure of trees using LiDAR's 3D measurement abilities serving as height proxies. While studies making use of deep learning algorithms have shown promising performances for the accurate mapping of canopy heights, they have limitations due to the type of architectures and loss functions employed. Moreover, mapping canopy heights over tropical forests remains poorly studied, and the accurate height estimation of tall canopies is a challenge due to signal saturation from optical and radar sensors, persistent cloud covers and sometimes limited penetration capabilities of LiDAR instruments. In this study, we map heights at 10 m resolution across the diverse landscape of Ghana with a new vision transformer (ViT) model optimized concurrently with a classification (discrete) and a regression (continuous) loss function. This model achieves better accuracy than previously used convolutional based approaches (ConvNets) optimized with only a continuous loss function. The ViT model results show that our proposed discrete/continuous loss formulation significantly increases the sensitivity for very tall trees (i.e., > 35m), for which other approaches show saturation effects. The height maps generated by the ViT also have better ground sampling distance and better sensitivity to sparse vegetation, in comparison to a convolutional model. Our ViT model has a RMSE of 3.12m in comparison to a reference dataset while the ConvNet model has a RMSE of 4.3m. The height map of Ghana being generated from free and open access remotely sensed data with sentinel-2 and Sentinel-1 images as predictors and GEDI height measurements, has the potential to be used globally. The ViT approach is data agnostic so that height maps at even higher resolutions could be obtained using higher resolution targets and predictors.

## 1. Introduction

Precise measurements of canopy height are essential to estimate important forest biophysical parameters, including above-ground biomass, carbon stock density, and habitat quality (Alexander et al., 2018; de Thoisy et al., 2016; Goetz et al., 2010). Canopy height also plays a critical role in ecosystem processes, such as carbon storage, forest productivity, and biodiversity (Lefsky et al., 2002; Zhang et al., 2016). Additionally, multi-temporal monitoring of canopy height is crucial for assessing forest disturbance regimes, dynamics, and associated deforestation and degradation (Pourshamsi et al., 2018), making it essential for guiding forest management and conservation efforts, carbon emissions reduction, and planning afforestation programs (Chen et al., 2016; Hese et al., 2005; Miles and Kapos, 2008). As a consequence, precise estimation of forest canopy height on a large scale has emerged as a pressing concern in current global carbon balance research (Giri et al., 2011).

Advancements in remote sensing (RS) technology has enabled the collection of an unprecedented amount of diversified data on forest structure and composition, both spatially and temporally, and has allowed researchers to estimate forest characteristics (e.g. canopy height, carbon content, tree cover) at an unprecedented resolutions ranging from 30 m products (Li et al., 2020; Potapov et al., 2021), to 10 m (Lang et al., 2022; Schwartz et al., 2022), down to the level of individual trees (Li et al., 2023; Mugabowindekwe et al., 2023; Reiner et al., 2023). Despite the differences in the spatial dimension of the different approaches, they all rely on the same types of remotely sensed information, namely LiDAR, radar, and multispectral (optical) observations.

LiDAR, an instrument mounted on aerial vehicles (aerial laser scanning, ALS), or operating from the Earth's orbit (spaceborne), emits laser pulses in the nadir direction and is able to accurately measure the time difference between transmitted emissions and their echoed returns. These returns, collected as 3D point clouds or as waveforms, have been proven to be the most accurate remote sensing way to estimate forest canopy heights (Alexander et al., 2018; Lefsky et al., 2005). Nonetheless, LiDAR's ground coverage remains limited to small areas due to operational costs of ALS campaigns, or technological limitations of spaceborne systems, that currently can only sparsely sample the Earth's surface (Dubayah et al., 2020; Neuenschwander et al., 2022). Optical (e.g. Sentinel-2, Landsat-8, PlanetScope) and radar (e,g. Sentinel-1, ALOS-2, PALSAR-2) systems on the other hand have repeated global coverage, with sampling resolutions ranging from high (e.g. 10-30 m) to very high (e.g. 3 m or less), with some missions providing free and open access to images. However, optical and radar systems do not have the ability to directly measure forest vertical structures. Indeed, optical systems only measure reflected sunlight from the Earth's surface. Over vegetation, this reflectance has shown correlation to the green leaf area (Turner et al., 1999), forest and vegetation structures (Woodcock et al., 1994), and even above ground biomass (Barbier et al., 2010; Couteron et al., 2005; Ploton et al., 2013). At high to medium image resolutions, the sensitivity of optical sensors, however, is limited to medium biomass levels due to sensor saturation (150-200 Mg/ha; Lu et al., 2012; Ploton et al., 2013). Side-looking radar, an active sensor that transmits electromagnetic waves at a given wavelength and measures the return power by the surface (backscatter), can penetrate through vegetation, with the depth of penetration increasing with longer wavelengths (Le Toan et al., 2011). As such, radar backscatter, which is a function of the geometric and dielectric properties of the volume of the scatterers in forests (e.g. leaves, branches, trunks) has been related to AGB (Baghdadi et al., 2015; Mitchard et al., 2012), and vegetation optical depth (Frappart et al., 2020; Wigneron et al., 2021). A saturation effect (i.e., decorrelation between the radar backscatter and AGB, even at L-band) has been reported (Imhoff, 1995; Sandberg et al., 2011).

To overcome the limitations of passive optical and radar systems' saturation and sparse coverage of LiDAR systems, and to ensure continuous canopy height mapping, current approaches rely on non-linear mathematical models, calibrated using machine learning algorithms. These algorithms learn a function that relates information acquired from the spatially continuous data (i.e. radar and optical data) to forest vertical structures (i.e. heights proxied by LiDAR measurements; Lang et al., 2022; Li et al., 2020; Morin et al., 2022; Ngo et al., 2023; Potapov et al., 2021; Schwartz et al., 2022). Machine learning algorithms can be classified into two major broad groups: conventional machine learning algorithms (ML), such as random forests (RF), or XGBoost, and deep learning algorithms (DL), which encompass the neural network-based family of algorithms. DL algorithms, in their broader definition, refer to a series of many connected layers that create a network were data passes through, from one layer to the next. While ML algorithms are easier to implement and are computationally less intensive than DL algorithms, they require structured data to be trained on. In essence, in the context of canopy height estimation, a pre-processing stage to extract relative vegetation indices from radar and optical data is required (Li et al., 2020; Morin et al., 2022; Ngo et al., 2023; Potapov et al., 2021). In addition, ML algorithms are unable to learn from the spatial context of the observation at the pixel scale. This, coupled with signal saturation from optical and radar imagery, severely impacts their ability to provide accurate height estimates of tall trees (Potapov et al., 2021). In contrast, DL algorithms extract relevant features from raw data as it passes through the network, alleviating the need for hand-crafted features, and offer immense flexibility in network architecture and loss function optimization. While computationally more expensive and harder to implement, DL algorithms can create tailored models that are better suited to learn the task than ML algorithms (Fayad et al., 2021b).

Current state-of-the-art approaches for large scale mapping of canopy heights all rely on the convolutional operation, which applies a set of learnable filters or kernels to the input image to produce a set of output feature maps. These convolutional layers extract specific features by sliding over the input image and performing element-wise multiplication and summation operations. By applying a series of convolutional layers, the DL network, also known as a ConvNet, can learn a hierarchy of features, from simple edges and textures to complex shapes and objects. For example, Lang et al., 2022 used an ensemble model of simple ConvNets to produce a global heightmap at 10 m pixel spacing using Sentinel-2 images as predictors and GEDI's $RH_{98}$ as height proxies. Schwartz et al., 2022 leveraged both Sentinel-1 and Sentinel-2 images for the estimation of canopy heights (GEDI's $RH_{98}$ as proxy) in the Landes forest in France using a ConvNet encoder-decoder architecture and showed higher accuracy than Lang et al., 2022. Recently, Liu et al., 2023 used two encoder-decoder ConvNets (one to classify between tree/no tree and the other for regression) to create a 3 m map of canopy heights across europe from PlanetScope images.

While such approaches show great potential for the large-scale mapping of canopy heights, they all rely on ConvNet based architectures, which are limited in their ability to capture long-range dependencies between pixels in an image (Yuan et al., 2021). Since convolutions operate on local neighborhoods, they cannot process information from the entire image in a single step, resulting in a lack of global context and a reduction in accuracy, especially when the input images contain complex relationships between pixels. Additionally, traditional ConvNets apply a single filter to the entire image, which results in a lack of expressiveness. As such, one possible limitation of the ConvNets approaches for canopy height mapping is their use of down-sampling strategies to increase the receptive field, which results in a loss of resolution and granularity of features in the deeper stages of the model that can be hard to recover in the decoder (Ranftl et al., 2021). To overcome this limitation, several techniques were introduced, such as the use of skip connections (Ronneberger et al., 2015), dilated convolutions (Yu and Koltun, 2016) that are meant to increase the receptive field of ConvNets, or more recently, the deployment of attention mechanisms at various stages of the network to enable selective focus on important features and better capture long-range dependencies (Chen et al., 2022, 2021; Oktay et al., 2018). While these techniques have proven to significantly improve prediction quality, ConvNet based models are still hindered by the limitations of the convolutional operation itself within the encoder. As such, a ConvNet based canopy height model might produce outputs with smooth details even though the input and output resolutions are the same.

Another possible limitation of current state-of-art models for the canopy height modeling is related to the used loss functions for model optimization. As we are dealing with regression tasks (i.e., canopy height estimation) current approaches rely on losses such as the L1 (MAE) and L2 (MSE), or a variation of those such as the Huber loss (Huber, 1964). However, these losses might not be suitable when dealing with a complex regression problem with a multi-modal or skewed distribution of targets, as is the case with height in heterogeneous land covers. In fact, such losses tend to minimize the average loss across all modes, resulting in a fit that cannot represent the extreme cases, such as tall trees. In a multi-modal scenario, they may produce a fit that satisfies neither mode (Mousavian et al., 2016). Therefore, canopy height models might fail to generalize height measurements over tall forests by employing a loss function that directly regresses the heights.

The main objectives of this study are therefore three-fold. First, to mitigate some of the limitations of ConvNet based models we will employ and evaluate a more recent family of deep learning models based on the transformer architecture. Transformer based models have become the standard in natural language processing tasks (Yuan et al., 2021). More recently, vision transformers (ViTs) have adapted the transformer architecture for computer vision tasks, demonstrating qualities such as the ability to capture long-range dependencies and a global receptive field (Vaswani et al., 2021). Moreover, a ViT based encoder could be an interesting alternative to a ConvNet one as it eliminates the need for explicit down-sampling operations. Instead, ViTs by design, maintain a consistent dimensional representation throughout the processing stages. However, while they have been successful with large datasets, they tend to perform worse than similarly sized ConvNets when trained on smaller amounts of data. This may be due to the lack of desirable properties in the ViT architecture that are inherent to ConvNets, like an inductive bias to focus on nearby image elements, making them less effective on smaller datasets (Touvron et al., 2021). As such, our second objective is to propose an approach for training vision transformers through a novel teacher/student learning paradigm which alleviates the need for additional data sources.

Our final objective is related to the underestimation of the taller trees due, in part, to the employed loss function. To address this issue, a more complex loss function will be proposed and is based on a discrete/continuous formulation, where the target values are classified into bins and the model is tasked to simultaneously estimate the target value and predict the corresponding class. Similar approaches have demonstrated better accuracy for regression problems (Kundu et al., 2018; Massa et al., 2016; Mousavian et al., 2016).

To this end, several DL based models as well as different loss functions will be assessed on their ability to estimate the heights across different land covers in an end-to-end matter. The different models we will evaluate will make no assumptions about the underlying forest cover or lack thereof, hence should perform well across a wide variety of land cover types. The models will be evaluated over the republic of Ghana. Ghana is an interesting case study due to its very diverse land cover, ranging from savannas to dense tropical forests. Hence, it should provide valuable insights on the abilities, as well as limitations, of our proposed approaches on the estimation of heights in different scenarios, ranging from sparsely vegetated areas in savannas, to the estimation of heights of tall trees within Ghana's tropical forests.

Finally, while our approach is data agnostic in terms of input and calibration data, we will make use of free and open access remote sensing data that have an extensive coverage both spatially and temporally, and that will remain operational in the near future.

The paper is organized as follows: the study site and the datasets are introduced in section 2. The methodology is presented in section 3. Section 4 describes the experimental settings, model parameters, and the results. Finally, the discussion and conclusions are presented in section 5.

## 2. Study area and datasets

*2.1. Study area*

The study area is Ghana, a country in West Africa with a total surface area of 238,535 km$^2$ (Fig. 1). It is bordered by Côte d'Ivoire to the west, Burkina Faso to the north, and Togo to the east, while its southern coast is bordered by the Gulf of Guinea. Ghana's terrain is characterized by low-lying coastal plains that rise to a central plateau. The terrain slopes gently from north to south, and the highest point is Mount Afadjato, which has an elevation of 885 meters.

Ghana has a tropical climate, with a rainy season from April to July and a dry season from November to March. The average temperature throughout the year is around 27°C, and the country experiences a relatively high level of rainfall, with an average of around 1500mm/year. The Koppen-Geiger classification (Peel et al., 2007) categorizes Ghana's climate as Aw, which is a tropical savanna climate.

Ghana's landscape is diverse, including forests, savannas, and wetlands. Forested areas, mostly located in the southern and central regions of the country, cover around 50,000 Km$^2$ of the total surface area. These forests are home to various tree species, such as the Ghanaian mahogany, the African teak, and the iroko. The forests in Ghana are classified into tropical rainforests, and mangroves. Tropical rainforests are the most abundant type, covering around 40% of the country's forested area. Mangroves, located along the coast, also cover around 30% of the country's forested area. Finally, the remaining 30% of trees are found in savannas, mostly located in the northern and north-east regions of Ghana (Fig. 1) and characterized by grasslands and scattered trees.

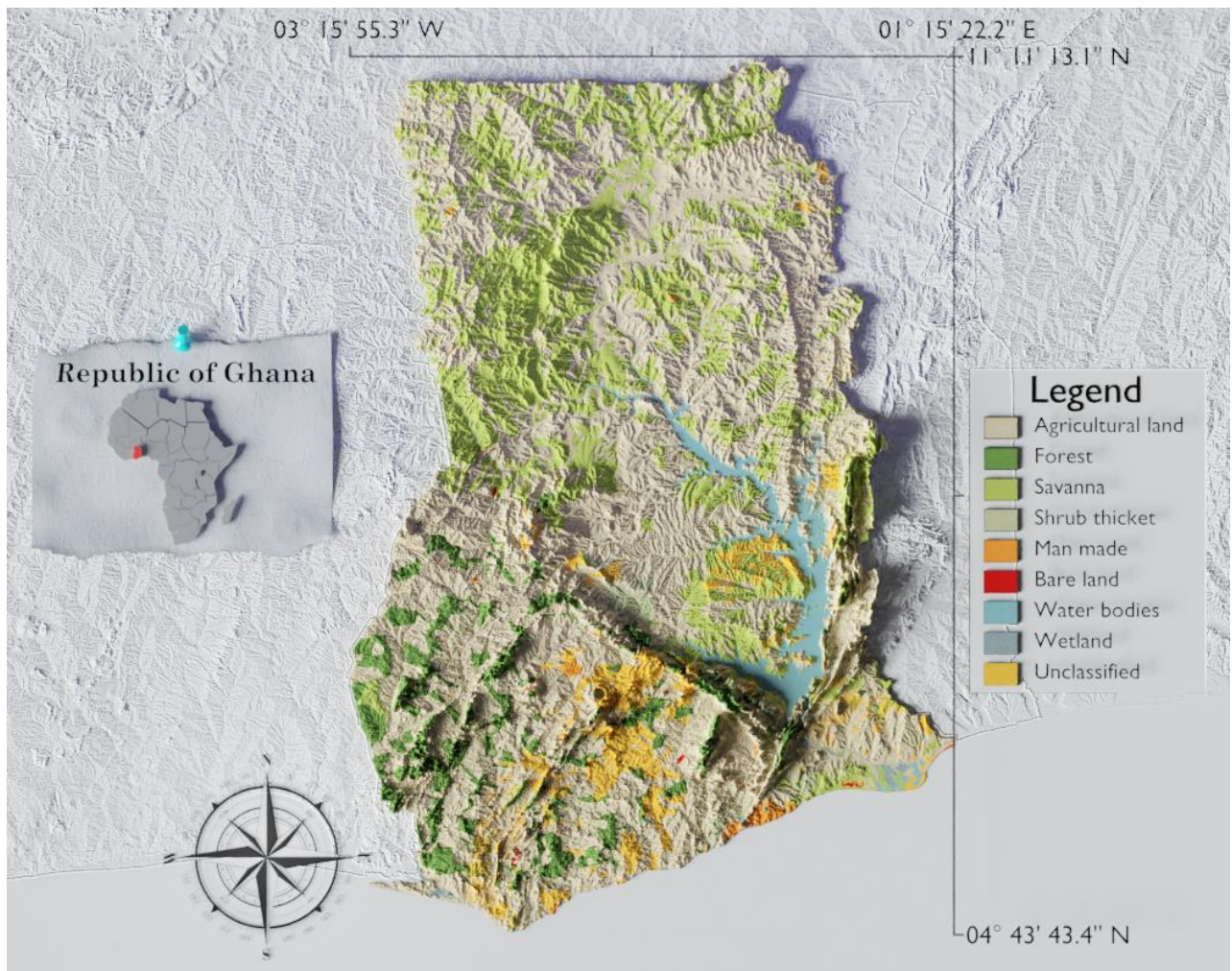

Figure 1. Location of the study area, the Republic of Ghana, and the different land cover classes (Source: https://geog.sdsu.edu/Research/Projects/IPC/research/ids.html).

*2.2. Datasets*

*2.2.1. Sentinel-2 Data*

We used the Sentinel-2 (S2) surface reflectance data (Level-2A) provided by Google Earth Engine (GEE). The Level-2A product (S2-L2A) comprises bands with a spatial resolution of 10m (Blue, Green, Red, and NIR bands) and 20m (Red Edge 1, Red Edge 2, Red Edge 3, Red Edge 4, SWIR1, and SWIR2 bands). For our study, all bands were upscaled with cubic interpolation to obtain bands with a 10m ground sampling distance (GSD). Next, S2 cloudy pixels were filtered out using the S2 Cloud Probability dataset provided by SentinelHub in GEE. After cloud removal, we created a mosaic using all cloud-filtered S2 images acquired between January 2019 and March 2020 by calculating the median of all time-series images over a selected pixel. Mosaicking has the potential to mitigate the effect of atmospheric conditions between time-series acquisitions that can lead to large differences between images. The choice to use 15 months of acquisitions is due to the persistent cloud cover in the southern part of Ghana as creating a composite from fewer acquisitions resulted in some areas with artifacts or missing pixels.

*2.2.2. Sentinel-1 Data*

SAR images were obtained from the Sentinel-1A (S1A) and Sentinel-1B (S1B) satellite constellation operating at the C-band (frequency of 5.405 GHz corresponding to a wavelength ~6 cm). S1 images, also provided by GEE, are calibrated using the Sentinel SNAP toolbox, developed by the European Space Agency (ESA). The calibration aims to convert the image pixel values into backscattering coefficients (σ°) in linear units, while the geometric correction was used to ortho-rectify the SAR images using the 30 m Shuttle Radar Topography Mission (SRTM) Digital Elevation Model (DEM). S1 SAR images were acquired in the Interferometric Wide swath mode (IW) in both VV and VH polarizations. These images were generated from the high-resolution Level-1 ground range detected (GRD) product, which were acquired at a spatial resolution of 20m × 22m (range × azimuth), and then resampled to produce the final product at 10m × 10m GSD. The radiometric accuracy of S1 SAR backscattering coefficients is approximately 0.70 dB (3σ) for the VV polarization and 1.0 dB (3σ) for the VH polarization (Schwerdt et al., 2017). Finally, the backscattering coefficients $\sigma_\theta^0$ acquired at different incidence angles θ were normalized to a common reference incidence angle set to $\sigma_{ref} = 40°$ using the square cosine correction equation (Baghdadi et al., 2001; Topouzelis et al., 2016):

$$\sigma_{ref}^0 = \frac{\sigma_\theta^0 \cos^2(\theta_{ref})}{\cos^2(\theta)} \qquad 1$$

As all active microwave systems, C-band S1 is sensitive to different physical phenomena that occur over a given landscape. Over bare and sparsely vegetated fields, soil moisture (SM) has a dominant influence on the C-band backscatter values (Baghdadi et al., 2017, 2008). To reduce the impact of SM, only images that were acquired in the absence of any rainfall events and up to four days prior (Bazzi et al., 2021) were retained. We define a rainfall event as rainfall (from GPM: Global Precipitation Measurement, V6) of more than 40 mm over a period of four days. Next, a composite using all images was created by calculating the median of all available acquisitions over a given pixel.

*2.2.3. Gedi Data*

GEDI is a full-waveform LiDAR sensor onboard the International Space Station (ISS) that has been in operation between latitudes of 51.6° N and 51.6° S since 2019. It is the first spaceborne lidar instrument specifically optimized to measure the vertical structure of vegetations (Dubayah et al., 2020). GEDI is equipped with three lasers emitting near-infrared (1064 nm) light. Two of the lasers are full power lasers, while the other full power laser is split into two beams designated the coverage lasers, for a total of four beams. Each beam is then optically dithered across-track resulting in eight ground tracks (four full power and four cover tracks) spaced 600 m on the ground. Shots have an average footprint of 25 m in diameter and are separated by 60 m along-track. The instrument can be rotated by up to 6°, allowing the lasers to be pointed up to 40 km on either side of the ISS ground track. This ability is used to sample the Earth's land surface as completely as possible, but can also affect the waveforms due to measurements made at angles farther than nadir (Fayad et al., 2020). In this study, we used GEDI's dataset Level 1B (L1B), Level 2A (L2A), and Level 2B (L2B) from April 2019 to December 2021. These datasets are published by the Land Processes Distributed Active Archive Center (LP DAAC). The L1B data product contains, among others, the geolocated waveforms. The L2A data product provides metrics relating to the vertical structure of vegetation from the processed waveforms using six possible settings groups (henceforth referred to as algorithms) of differing thresholds to satisfy a variety of acquisition scenarios (Dubayah et al., 2020). Moreover, in the L2A data product, a variable named 'selected_algorithm' corresponds to the algorithm with the lowest (in terms of elevation) non-noise ground return (henceforth referred to as the "lowest mode"). Therefore, from the L1B we extracted the waveforms, and from the L2A, and L2B data products we extracted the following variables derived from the processing algorithm identified by the selected_algorithm variable:

- The latitude, longitude, and elevation of the lowest mode.
- The latitude, longitude, and elevation of the instrument.
- The number of detected modes (num_detectedmodes).
- The acquisition date and time of the shot.
- The beam sensitivity, that is, the probability of detecting the ground below a given canopy cover 90% of the time with a 5% chance of a false positive (Hancock et al., 2019).
- The viewing angle (VA) at acquisition time (Fayad et al., 2022).
- The signal-to-noise ratio (SNR) (Fayad et al., 2022).
- The relative height at 98% ($RH_{98}$) of returned energy as a proxy of canopy height (Blair et al., 1999).

- The elevation of the lowest mode (ELM) and the elevation of the highest return (EHR), which represents the canopy top.
- Canopy cover

Over Ghana, 27,096,322 shots were acquired during the period between April 2019 and November 2021. However, not all of these shots are viable due to unfavorable atmospheric conditions (e.g., clouds) that affected them, or other instrument related errors. Therefore, we removed all waveforms that met any of the following criteria:

- Shots without any detected modes (num_detectedmodes = 0). These shots were noisy signals with a signal-to-noise ratio (SNR) of 0 on a linear scale.
- Shots with a SNR < 12db or VA > 5°.
- Shots with beam sensitivity lower than 95%.
- Shots with an absolute difference between the elevation of the lowest mode (ELM) and the corresponding SRTM DEM of more than 75 m ($|ELM - SRTM| > 75m|$).
- Incomplete waveforms, as in waveforms with an insufficient number of bins. These waveforms are detected by their search_end variable (end location of the usable part of the waveform) that is near the total number of bins in a waveform "rx_sample_count" ($rx\_sample\_count - search\_end \leq 1\ bin$).
- Observations where the absolute difference between GEDI's canopy cover and S2 derived NDVI (upsampled to 30 m) was greater than 1.5 standard deviation (1.5σ) from the mean (Liang et al., 2023). This filter was used to remove shots that potentially have high geolocation errors.

After Filtering, 43% of the shots were retained over Ghana. Fig. 2 shows the distribution of $RH_{98}$ over the entire study site.

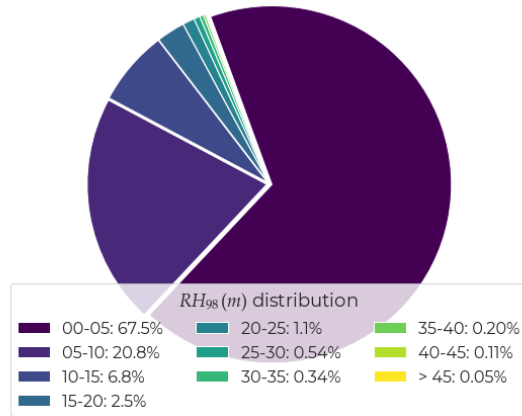

Figure 2. Distribution of GEDI's $RH_{98}$ (m) acquisitions over Ghana.

*2.3. ALS Data*

An airborne laser scanning dataset provided by GlobALS (Stereńczak et al., 2020), acquired over an area of ~10km$^2$ of tropical forests in southwestern Ghana in 2012 was used for validation (Fig. 3). The ALS dataset was acquired using Optech gemini which is a wide-area, high-altitude airborne discrete lidar mapping system with a point density of 10/m$^2$. From the point cloud dataset, two rasters were generated: a digital elevation model (DEM) representing ground elevations, and a digital surface model (DSM) representing the height of the top of the canopy. The canopy height model (CHM) was generated by subtracting the DSM from DEM at 1m resolution.

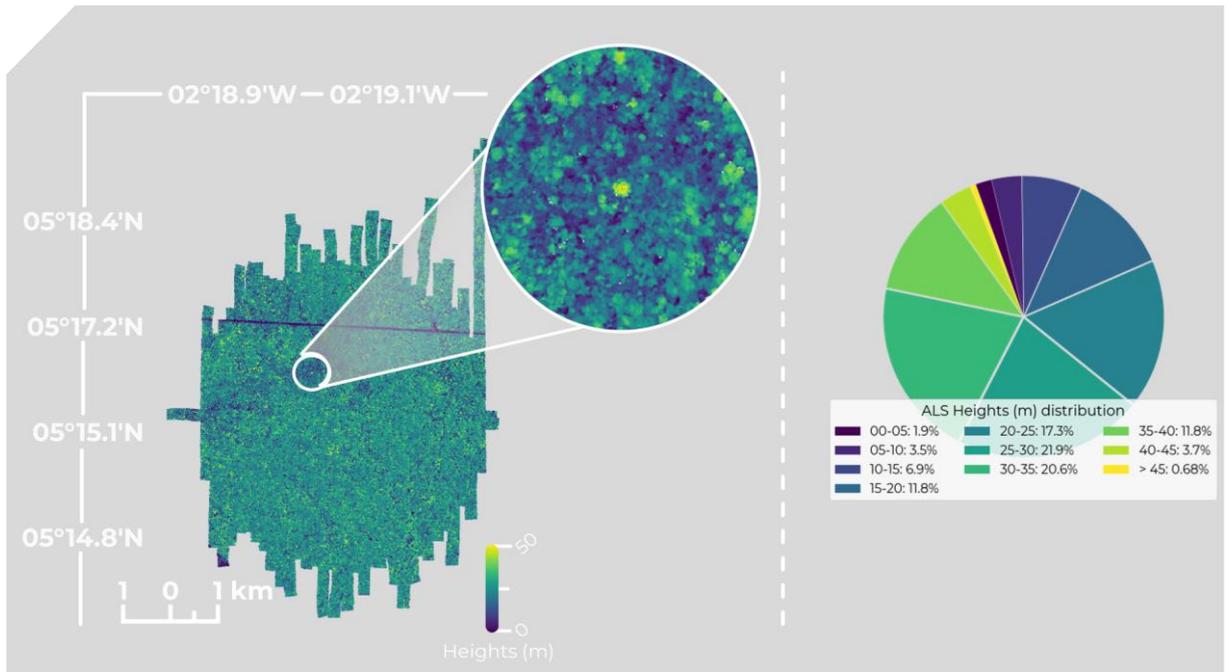

Figure 3. The ALS dataset and the distribution of the heights.

*2.4. Stereo 3D reconstruction from Pleiades imagery*

Pleiades is a constellation of optical satellites owned and operated by Airbus. Pleiades acquires very high resolution (VHR) images in panchromatic, visible, and infra-red bands at 0.7m (panchromatic) and 2.8m (multispectral) resolutions and a geolocation accuracy better than 5m. To derive a height map from Pleiades VHR acquisitions, we adapted the stereo 3D height reconstruction technique from the study of De Franchis et al., 2014, and a stereoscopic Pleiades acquisition over a sparse vegetated area in 2016. The height map is obtained by first creating a digital surface model (DSM) at 0.7m resolution from point-cloud 3D reconstruction based on the two Pleiades images. Then a cloth simulation algorithm was applied to select the ground points from the 3D point Cloud (Zhang et al., 2016), which were then interpolated by a Laplace operator (Ousguine et al., 2016) to obtain a digital elevation model (DEM). Finally, a canopy height model (CHM) was obtained by subtracting the DEM from the DSM. Fig. 4 shows the 34Km$^2$ canopy height map (hereafter referred to as VHR-HM) derived using the pre-mentioned approach.

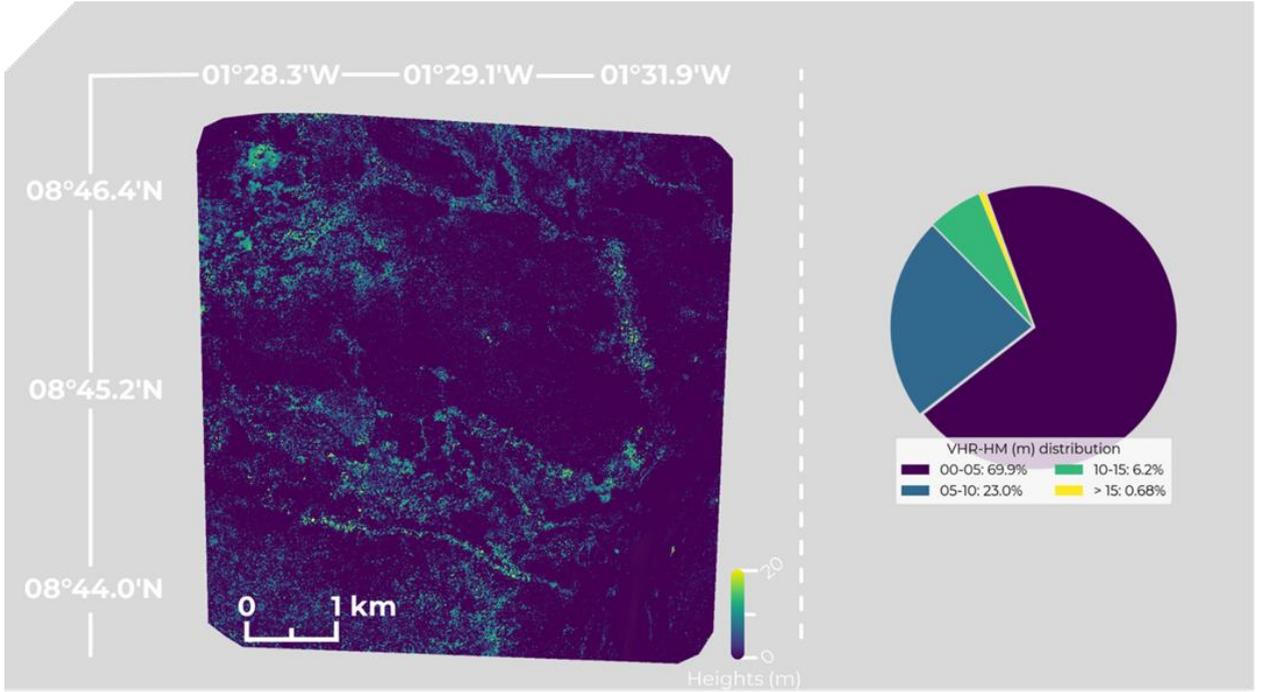

Figure 4. Illustration of a Stereo 3D reconstruction of canopy height from Pleiades imagery (0.7m resolution) over a forested region of Ghana and associated distribution.

## 3. Methodology

In this study, we aim to evaluate the performance of various deep learning architectures for the generation of wall-to-wall land cover height maps. The models will be trained using high-resolution (10m) optical and/or radar imagery as input and calibrated on GEDI height estimates serving as proxy of ground truth heights. This section will detail model architectures as well as the loss functions used to optimize each model.

### 3.1. ConvNet based models

Our first type of models is inspired by the encoder-decoder architecture from the study of Chen et al., 2022, which reflects current advances on the U-Net architecture. The U-Net and its variants are a type of ConvNets that are commonly used for image segmentation, dense prediction and other tasks involving image data. First introduced by Ronneberger et al., 2015 for the segmentation of biomedical images, it has since been successfully deployed for various applications across multiple disciplines, and has shown great potential for the estimation of canopy heights (Li et al., 2023; Schwartz et al., 2022). The primary function of a U-Net is to learn the mapping between the input and output images by learning the features within an input image at different scales. It consists of a compressive path (encoder) to extract features from the input image and an expansive path (decoder) to reconstruct the output image from the encoded features. This section presents the different parts of our U-Net based models.

#### 3.1.1. Encoder path

The encoder path used consists of four Convolutional based Encoder Blocks ($CEB_i$, Fig. A.20) that extract increasingly abstract and high-level features from the input image while reducing its spatial resolution at each stage. Each $CEB_i$ comprises two convolutional operations, followed by a Batch Normalization (Batch Norm) layer and a Leaky ReLU activation function. Leaky ReLU was chosen over the standard ReLU activation function in each block, as it overcomes some of its limitations. Unlike ReLU, which sets all negative input values to zero, Leaky ReLU introduces a small slope for negative values (0.01), enabling the network to learn even for negative inputs. This property ensures that no neuron becomes entirely inactive. Finally, to reduce the spatial dimension, a strided 2×2 convolutional layer was used instead of a pooling layer (e.g. max-pooling) to preserve information in the feature maps. The use of strides rather than max pooling was preferred as max pooling retains only the maximum values in a given area, which could result in information loss. Each encoder block with input of size: width (W), height (H), and channels (C) results in a feature map of size $\frac{H}{2} \times \frac{W}{2} \times 2C$:

$$CEB_i: \mathbb{R}^{W \times H \times C} \rightarrow \mathbb{R}^{\frac{W}{2} \times \frac{H}{2} \times 2 \cdot C} \qquad 2$$

Optical and radar images will be jointly used to produce wall-to-wall height maps. Given that the two data sources represent distinct physical measurements over a scene, our U-Net based implementation will use two separate encoder paths, one for each data source. By having separate encoders, the network can learn to extract relevant features and representations independently for each modality, rather than trying to extract common features for all modalities (Kuga et al., 2017). This approach can lead to a more focused representation of the data, as each encoder can specialize in processing its own modality (Stahlschmidt et al., 2022). The outputs of both encoder paths are then concatenated channel-wise and fed into a self-aware attention (SAA) module to effectively learn the non-local interactions among encoder features. The specific implementation of the SAA module can be found in the study of Chen et al., 2022.

*3.1.2. Decoder path*

Following the SAA module, which represents the bottleneck of the model, is the decoding path, which uses the encoded features to reconstruct the output image. It does so by upsampling the feature maps and refining them with convolutional layers. The decoding path used in this study consists of four Convolutional-based Decoder Blocks ($CDB_i$, Fig. B.21) that include upsampling layers, concatenation layers, and convolutional layers. The upsampling layers increase the spatial resolution of the image, while the concatenation layers connect the decoding path to the encoding path through skip connections, allowing the network to consider the features extracted in the encoding path as well (Ronneberger et al., 2015).

Each of the four decoder blocks start with a strided 2x2 transpose convolution operation (ConvT) which effectively doubles the spatial dimension of the input feature map. ConvT allows the network to learn a mapping from low-resolution to high-resolution feature maps, whereas bilinear interpolation is a fixed and non-learnable operation. This means that the network can use its knowledge of the data to generate more meaningful high-resolution feature maps, rather than simply stretching the low-resolution maps as in bilinear interpolation. The decoder block scales the input feature maps of size W×H with C channels to a size of: $2W \times 2H \times \frac{C}{2}$:

$$CDB_i: \mathbb{R}^{W \times H \times C} \rightarrow \mathbb{R}^{2 \cdot W \times \cdot H \times \frac{C}{2}} \qquad 3$$

*3.1.3. Output head and model optimization*

Finally, an output head is connected to the decoder path. The purpose of the output head is to map the high-level features learned by the network to the desired output, which are the estimated heights. Given that the task is a regression problem, the implemented output head consists of a single convolution operation with a kernel size of 1 and a stride of 1. This operation reduces the number of channels from C channels to a single channel (i.e., the height estimates), and is followed by a Softplus activation layer (SoftPlus is a smooth approximation of the ReLU function) to ensure that the output heights are always positive (Fig. 5.a). The full implementation of the network, referred to as 2MOU (MultiModal, single Output UNet), is presented in Fig. 6.

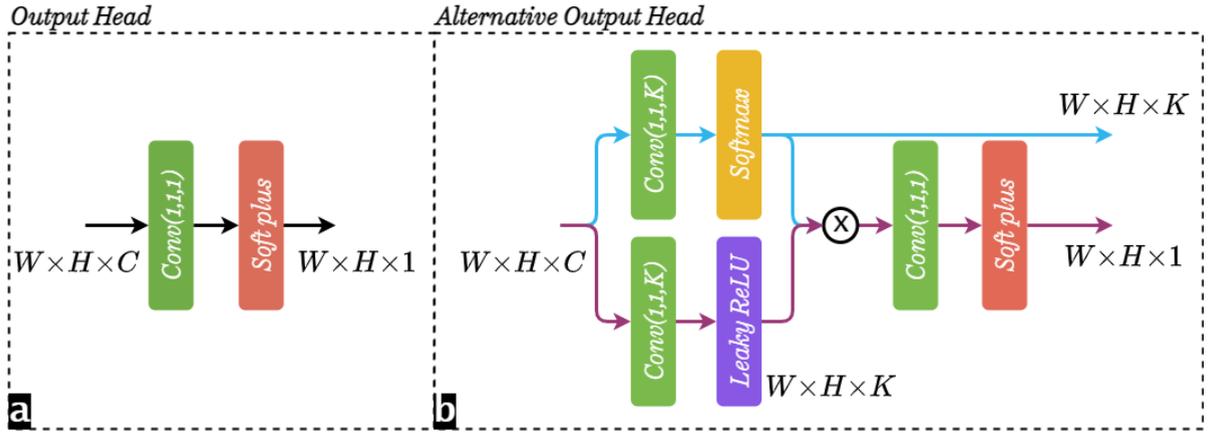

Figure 5. Proposed output heads for the height estimation models. *ConvT(K, S, F)* represents a convolution operation with a kernel of size K, stride S, and F filters. In (b) the blue path represents a classification sub-module and the purple path represents a regression sub-module.

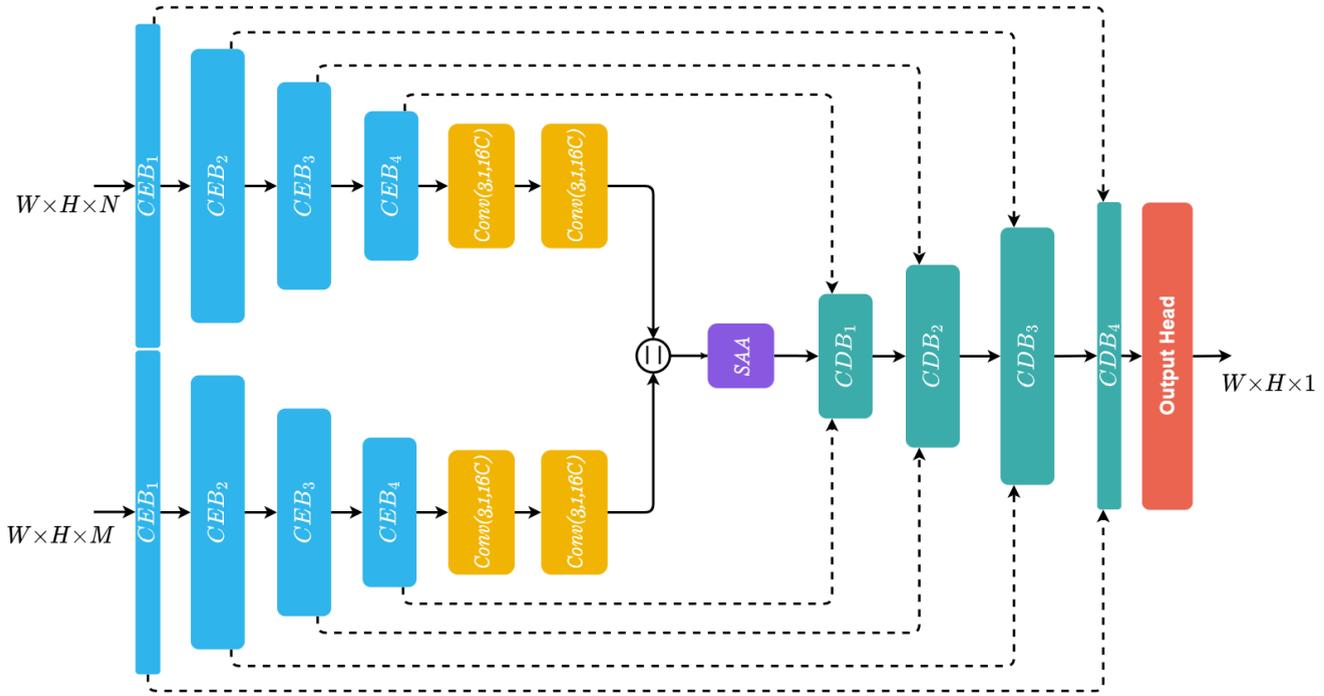

Figure 6. The architecture of the 2MOU model. $CEB_i$ represents an encoder block, $CDB_i$ represents a decoder block. Skip connections are represented by dashed lines. The *Output Head* represents the output block.

In terms of model optimization, as we are dealing with a regression problem, we adopt the Huber loss (Huber, 1964) and the loss function of the 2MOU model is defined as follows:

$$\ell_{2MOU}(Y,Y') = \ell_H(Y,Y') = \frac{1}{N}\sum_{i=1}^{N} \begin{cases} \frac{1}{2}(Y_i - Y_i')^2, & if\ |Y_i - Y_i'| < delta \\ delta * \left(|Y_i - Y_i'| - \frac{1}{2} \cdot delta\right), & otherwise \end{cases} \quad 4$$

Where $Y$ is the vector of target values to estimate and $Y'$ the vector of truth values, N is the number of samples, and delta is a hyperparameter which determines the threshold for the function to switch from quadratic loss to linear. In this study, delta was set to 3m. The choice to use the Huber loss is that it is more robust to outliers than the Mean Squared Error (MSE) loss and preserves some of the regularity in the loss surface. Unlike the MSE loss, which penalizes large errors quadratically, Huber loss linearly penalizes large errors beyond a certain threshold, making it less sensitive to outliers. Similarly, the Mean Absolute Error (MAE) loss also linearly penalizes errors but lacks the smooth transition in the penalization for large errors, as provided by the Huber loss. As such, Huber loss is a better compromise between robustness and regularity in comparison to the MSE and MAE loss functions.

*3.1.4. Alternative output head*

In many machine learning regression tasks, the most common regression losses used are the L1 (MAE) and L2 (MSE), or a variation of such losses such as the Huber loss. Nonetheless, as stated earlier, when dealing with non-normal distributions of the targets, such as the case of low number of GEDI acquisitions over tall trees in comparison to shorter ones, a model trained with such a family of losses may under- or over- estimate extreme values. Over Ghana, Fig. 2 shows that the majority of acquired GEDI measurements are less than 15 m (~95% of footprints). Therefore, we suspect that by regressing the targets directly, the model will fail to generalize GEDI measurements over tall forests. To mitigate this issue, we propose an alternative output head for the model presented in the previous section (Fig. 5.b). The modified output head will perform two tasks. The first task is to attempt to classify the heights into $K$ overlapping bins (classification sub-module), while the second task will output continuous height estimates (regression sub-module). This discrete/continuous formulation has shown better accuracies for regression problems (Kundu et al., 2018; Massa et al., 2016; Mousavian et al., 2016). In fact, classification over-parametrizes the problem, offering the network more flexibility to learn the task. Moreover, the classification sub-module could also be encouraged to pay more attention to underrepresented classes (i.e., tall trees) by weighing the contribution of each class to the total classification loss.

In terms of implementation, the modified output head (Fig. 5.b) will connect the output of $CDB_4$ (Fig. 6) of size $W \times H \times C$ to two separate convolutional layers with size $W \times H \times K$ (K is the number of the predefined overlapping height classes). To obtain height class probabilities a *softmax* activation function is used on one of the convolutional layer's outputs (Fig. 5.b, blue path). Finally, the outputs of the two convolutional layers of size $W \times H \times K$ are multiplied element wise and go through to another convolutional layer with an output of size $W \times H \times 1$ followed by a *Softplus* activation function, making the regression task (Fig. 5.b, purple path). As such, the classification path will act as a constraint on the regression sub-module by bounding its output. Fig. 5.b shows the full implementation of the alternative output head. This second implementation of the U-Net model will hereafter be referred to as 2MDU (MultiModal, Dual output UNet).

In terms of loss functions, as we are trying to optimize the model for both the classification and regression tasks, the total loss ($\ell_{CR}$) of the model is defined as:

$$\ell_{CR}(P,T,Y,Y') = \ell_{CE}(T,P) + \alpha \cdot \ell_R(Y,Y') \qquad 5$$

Where *P* and *T* are respectively the predicted and true probabilities for each class and for each target and truth value, $\ell_R$ is the regression loss, and α is a scaling parameter to balance both losses. $\ell_{CE}$ is equal to a weighted cross-entropy loss since we are dealing with an imbalanced dataset. Cross-entropy loss is a commonly used loss function for multi-class machine learning models. It measures the difference between the predicted probability distribution and the true distribution represented by the confidence labels. In a multi-class scenario, the cross-entropy loss calculates the average negative log likelihood of the correct class. The cross-entropy loss ($\ell_{CE}$) is defined as follows:

$$\ell_{CE}(P,T) = -\frac{1}{N}\sum_{i=1}^{N}\sum_{j=1}^{k} w_j \cdot t_{ij} \cdot \log(p_{ij}) \qquad 6$$

Where $t_{ij}$ and $p_{ij}$ are respectively the true and predicted probabilities of class $j$ for target value $i$, $w_j$ is the weight assigned to class $j$. $w_j$ is equal to the inverse ratio of GEDI acquisitions belonging to a certain class to the total number of GEDI acquisitions in a single batch.

For the regression loss ($\ell_R$), we will test two regression losses, the first one is the Huber loss (eq. 4). However, the residual (difference between predicted and true values) is not bounded in a Huber loss, with a magnitude that is linearly proportional to the difference between model estimates and the GEDI targets for large differences, and quadratically proportional to the difference for small differences. As model optimization progresses, estimated values that are close to high quality GEDI measurements will have little contribution during backpropagation, while estimates that are far from GEDI measurements, due for example to GEDI's geographical location errors, or instrument errors (e.g. limited laser penetration in densely vegetated forests) will produce a gradient with high magnitudes, and as a result, the model will adapt to these outliers and deteriorate its performance for the inliers. To reduce the effect of these outliers, we will replace the Huber loss by an adaptive and robust regression loss developed by Barron, 2017 (C.11). The 2MDU model optimized with the adaptive loss function will hereafter be referred to as α-2MDU.

### 3.2. ViT based model

Another proposed approach will be based on a hybrid ViT-ConvNet model specifically tailored for the continuous prediction of land cover heights. As we are dealing with a dense prediction problem, the proposed model is also based on an encoder-decoder architecture with a ViT serving as an encoder and a ConvNet will be used as a decoder. As we stated in the introduction, transformer based models require an order of magnitude more data than similarly sized ConvNets (Touvron et al., 2021). Moreover, ViT based models, unlike ConvNets which are designed to exploit the spatial structure of images, rely solely on the attention mechanism to extract features from images. This means that ViT based models need to learn how to attend to relevant parts of the image and how to combine information across different regions of the image. This requires a stronger inductive bias to guide the learning process. As such, ViTs usually require long pre-training times, before being fine-tuned to the required task (He et al., 2021). To counter the issue of data limitation, and long pre-training times, we get inspiration from the study of Touvron et al., 2021, and we couple the proposed architecture with two pre-trained UNet based models, and optimized the model in a teacher/student training paradigm.

This section introduces our hybrid ViT-ConvNet model, hereafter referred to as Hy-TeC. We first start by processing the input images to be used by the transformer, then we present our encoder and decoder architectures, finally, we present our training strategy.

#### 3.2.1. Image patching

Since ViTs use the original transformer architecture conceived for the processing of sequential data (e.g., sequence of words), fixed-sized input images with shape (width W, height H, and C channels) are first decomposed into a batch of $N$ non-overlapping patches or tokens, of shape P×P×C. Each token is then projected by a linear layer $L$ to a vector of size $D = P \cdot P \cdot K$. Next, to conserve positional information, a learnable positional encoding (Jiang et al., 2022) is applied to each projected token (hereafter referred to as embedded patches). Moreover, thanks to the global receptive field of ViTs which allows each pixel to attend to every other pixel within an image, we will separate the inputs to the encoder by their native resolution. In essence, bands with an original resolution of 10m will be concatenated channel-wise in input 1, while bands such as the NIR and SWIR bands from S2 that have a native resolution of 20m will be concatenated channel-wise in input 2. Therefore, the overall input to the model is the concatenated embedded patches from input 1 and input 2. The pre-processing of an image to be used by the ViT is shown in Fig. D.22.

#### 3.2.2. Encoder

The encoder is composed of multiple standard Transformer layers (Dosovitskiy et al., 2020). Each layer consists of a normalization layer (Norm), followed by a Multi-Head Self-Attention module (MHSA), another normalization layer, and finally a multi-layer perceptron (MLP). The input tokens fed into the encoder are transformed using $T$ transformer blocks into a new representation $t^i$, where $i$ is the output of the $i$-th transformer layer. In this study we used the ViT-Base proposed in (Dosovitskiy et al., 2020) which features 12 transformer blocks.

*3.2.3. Decoder*

The decoder layers are inspired by the decoder layers presented in the study of Ranftl et al., 2021. Also, as the encoders are fed using 2N embedded tokens of size D from input 1 (native 10 m resolution) and input 2 (native 20 m resolution), only the N transformed tokens from input 1 are used in the decoder layers. In essence, a reprojection block ($RB_i$, Fig. E.23) followed by a decoding block ($DB_i$, Fig. F.24) is responsible for reshaping the transformed input 1 tokens into feature maps of different resolutions. The first operation in the reprojection block is a spatial concatenation (SP) of the tokens of size $N \times D$ to an image-like feature map of size $P \times P \times L$:

$$SP = \mathbb{R}^{N \times D} \rightarrow \mathbb{R}^{P \times P \times L} \qquad 7$$

After the *SP* operation, the feature maps from four different transformer blocks are reprojected to four different resolutions with $\hat{L}$ channels:

$$RB_i = \begin{cases} \mathbb{R}^{P \times P \times L} \rightarrow \mathbb{R}^{\frac{P}{2} \times \frac{P}{2} \times \hat{L}}, if\ i = 1 \\ \mathbb{R}^{P \times P \times L} \rightarrow \mathbb{R}^{P \times P \times \hat{L}}, if\ i = 2 \\ \mathbb{R}^{P \times P \times L} \rightarrow \mathbb{R}^{2P \times 2P \times \hat{L}}, if\ i = 3 \\ \mathbb{R}^{P \times P \times L} \rightarrow \mathbb{R}^{4P \times 4P \times \hat{L}}, if\ i = 4 \end{cases} \qquad 8$$

Next, the four reprojected blocks are input to four decoder blocks (DB), consisting of two convolutions operations (Conv K=3, S=1) and a transpose convolution operation (ConvT K=4, S=4) to effectively quadruple the spatial dimension and reduce the number of channels by eight.

$$DB_i = \mathbb{R}^{\hat{H} \times \hat{W} \times \hat{D}} \rightarrow \mathbb{R}^{4 \cdot \hat{H} \times 4 \cdot \hat{W} \times \frac{\hat{D}}{8}}, i \in \{1, \dots, 4\} \qquad 9$$

Finally, to produce height estimates, we connect the same output head in Fig 5.b to the output of $DB_4$. Effectively making our Hy-TeC model a height classifier and regressor.

*3.2.4. Knowledge distillation*

Knowledge distillation (KD), introduced by Hinton et al., 2015, is a technique of transferring knowledge from a pre-trained teacher model to a smaller student model without incurring significant drops in performance. The success of KD can be attributed to the fact that deep learning models are highly complex, non-linear functions that can be difficult to optimize. By emulating the outputs of a pre-trained teacher model, the student model is able to inherit the knowledge and generalization abilities of the teacher model, thus simplifying the learning task. In general, KD in a classification problem is achieved through the softened cross entropy loss, which regards the teacher predictions as 'soft labels' containing knowledge of the correlation of the class labels (Hinton et al., 2015). ViT based models can benefit from KD as demonstrated in a recent work with transformer based models (Touvron et al., 2021). Indeed, Touvron et al., 2021 were able to train a ViT model on a reduced dataset with performances comparable to a ViT trained on large amounts of data. Nonetheless, KD works well for classification problems since the student models have access to the softened labels of the teacher model. In regression problems where the output space is continuous, the teacher model predicts values with the same characteristics as the ground truth data and an additional unknown error distribution. Moreover, since the output space is unbounded, a teacher model can provide false guidance to the student model. As such, KD is harder to achieve.

Here we propose a novel KD approach for the optimization of our proposed transformer-based model to forgo the need of additional training images. First, acting as teacher models are two single-encoder U-Nets (c.f. section 3.1) each pre-trained on a given modality (i.e., S1 and S2 images). The choice of estimating heights from the two modalities, while each being less accurate than a model using both modalities, can act as a filtering criterion of the data that we wish to transfer to the transformer-based model. Indeed, we assume that if estimates from two modalities are relatively similar, the probability of them representing the true height is higher. Next, since we cannot compute soft labels from the teacher models, which is a requirement for KD, we instead use the filtered output of the teacher models to influence the transformer-based model at lower resolutions, while leaving it to freely form its own representation at higher resolutions (i.e., 10m).

In terms of implementation, the output of the two teacher models at the highest resolution with size $H \times W$ will be downsampled bilinearly to fit the output resolution of the decoder blocks of the transformer-based model at lower resolutions. Next, the average value of the two heights estimates from the two teacher models (one model using S1 data, and the other using S2 data) is computed only if the percentage difference is lower than 10%. Finally, the model will need to optimize six loss functions. At the highest resolution (i.e., 10m), where the model outputs height estimates and height class labels, the two loss functions are identical to eq. 5. For the KD loss, the transformer-based model can output through the decoder blocks height estimates at resolutions of $\frac{W}{16} \times \frac{H}{16}$ (160 m), $\frac{W}{8} \times \frac{H}{8}$ (80 m), and $\frac{W}{4} \times \frac{H}{4}$ (40 m). These blocks will try to emulate the output of the teacher models at the same resolutions, and thus the associated loss functions will be identical to eq. 4. The choice not to use the loss function in eq. 5 for the lower resolution outputs, is simply to reduce model complexity. As such, the full loss function that the Hy-TeC model will try to minimize is defined as:

$$\ell_{HT}(H_{db1}, H'_{db1}, H_{db2}, H'_{db2}, H_{db3}, H'_{db3}, H_{db4}, H'_{db4}, HR_{db4}, HR'_{db4})$$
$$= \beta_1 \cdot \ell_H(H_{db1}, H'_{db1}) + \beta_2 \cdot \ell_H(H_{db2}, H'_{db2}) + \beta_3 \cdot \ell_H(H_{db3}, H'_{db3}) + \beta_4 \qquad 10$$
$$\cdot \left(\ell_{CE}(HR_{db4}, HR'_{db4}) + \alpha \cdot \ell_{RA}(H_{db4}, H'_{db4})\right)$$

Where $\ell_H$ is the Huber Loss, $\ell_{CE}$ is the cross entropy loss, $\ell_{RA}$ is the robust adaptive loss, $H_{dbi}, i \in \{1,2,3\}$ are the height estimates from $DBi, i \in \{1,2,3\}$, while $H'_{dbi}, i \in \{1,2,3\}$ are the height estimates from the teacher models. $H_{db4}$ and $H'_{db4}$ are respectively the height estimates from $DB4$ and GEDI. $HR_{db4}$ and $HR'_{db4}$ are respectively the height classes from $DB4$ and GEDI. $\beta_i, i \in \{1, ..., 4\}$ are scaling hyper-parameters that we set to 1 for $\beta_4$ and 0.7 for the rest. The full implementation of the Hy-TeC model is presented in Fig. 7.

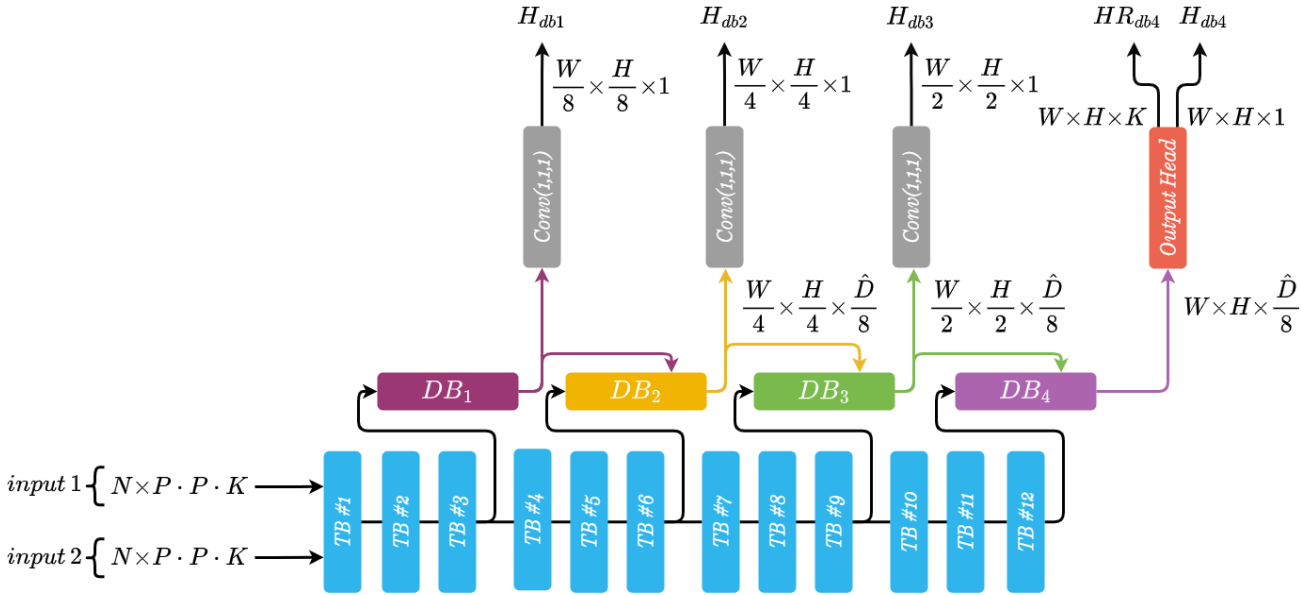

Figure 7. The implementation of the Hy-TeC model.

# 4. Experimental settings

The four proposed models (2MOU, 2MDU, α-2MDU, and Hy-TeC) will be evaluated on their capability to estimate canopy heights and heights across different land cover types over Ghana. We will first start by introducing the data requirements for the models, followed by the training strategy, model evaluation metrics, and finally the results. Model inputs, and the associated loss function of each model are presented in the following table (Table 1).

| Model designation | Inputs | Architecture | Loss function |
|---|---|---|---|
| 2MOU | S1 and S2 data | ConvNet | Huber loss |
| 2MDU | S1 and S2 data | ConvNet | Cross entropy + Huber loss |
| α-2MDU | S1 and S2 data | ConvNet | Cross entropy + Adaptive loss |
| Hy-TeC | S2 data | Hybrid (vision transformer encoder and ConvNet decoder) | Cross entropy + Adaptive loss |

Table 1. Summary of the four evaluated models, their input data, architecture, and the loss function used for optimizing each model.

*4.1. Sampling strategy*

Due to the unbalanced distribution of GEDI measurements over the study area, two sampling steps were defined to derive the datasets for the input image patches (i.e., S1, and S2) and the targets (i.e., GEDI measurements) that the different models will use in the training and validation stages. First, a regular 768×768 (i.e., 7680m×7680m) grid was overlaid over Ghana, where for each grid cell, the distribution of GEDI height measurements was computed at 5 m intervals. This resulted in 2559 cells containing at least 600 GEDI acquisitions. To guarantee that during training (resp. validation) the models are trained (reps. validated) across all height ranges, we further split the 2559 cells by the ratio of the available $RH_{98}$ within each grid cell, and selected the grid cells containing suitable patches of inputs and targets based on appropriate ratios for each height range (Table G.3).

Next, grid cells from each set were further split to 75% used for training and 25% used for validation. We also duplicated the under-represented sets n times (Table G.3) to better balance the training and validation datasets. Finally, during a training step (resp. validation), a 256×256 randomly defined bounding box was used to extract the necessary S1 or S2 image patches used as model inputs, and the corresponding $RH_{98}$ as the targets. Even though during a training step (resp. validation), the same training (resp. validation) grid-cell could be selected multiple times, overfitting is controlled by image augmentation techniques (random crop and flip).

*4.2. Model inputs and training*

For the UNet based models, the selected 256×256×10 (W×H×C1) S2 images and the 256×256×2 (W×H×C2) S1 images were fed to the models in batches of 12 without any further modification. For the transformer-based model (i.e., Hy-TeC), only S2 bands were considered as inputs to the encoder due to the relationships between SAR backscatter and forest biomass that could be disturbed by several parameters such as forest structure, soil and vegetation moisture, and understory rebound. As such, input 1 is composed of four S2 bands (R, G, B and NIR), while input 2 is composed of six S2 bands (red edge 1, 2, 3, and 4 and SWIR 1 and 2). The 256×256 images with B bands for input 1 and input 2 (B = 4 for input 1 and 6 for input 2) were first divided into 256×16×16×B (N×P×P×B) non-overlapping patches and each patch flattened to a 1D vector of size 16×16×B. We then map each flattened patch vector to a lower-dimensional space by applying a linear embedding of size 1536, representing the tokens. Finally, we add the positional encoding to the projected tokens and concatenate the embeddings from input 1 and input 2 to form a 2D matrix of size 512×1536 (256×1536 from input 1 and 256×1536 from input 2), which are then fed in batches of 12 to the model. In terms of optimization, a Stochastic Gradient Descent (SGD) optimizer is used for the UNet based models with an initial learning rate of 1e-2 and a total batch size of 12 for a maximum of 250 epochs, with a cosine learning rate decay scheduler. The transformer based model on the other hand is trained with the AdamW (Loshchilov and Hutter, 2019) optimizer with a linear warmup scheduler that gradually increases the learning rate from 1e-6 to 1e-4 over 20 epochs, after which a cosine learning rate decay scheduler is used. The transformer model is trained with a batch size of 12, with an image patch size of 16×16 for a maximum of 250 epochs including the warmup phase.

*4.3. Model performance evaluation*

The models will be assessed on four criteria: (1) Their generalization capabilities on the calibration data (i.e., GEDI acquisitions), (2) Their accuracy in comparison to the ALS and VHR-HM maps, (3) Model output resolution, and (4) map qualities.

Comparison to the set-aside GEDI $RH_{98}$ estimates as well as the ALS dataset will be conducted using standard summary statistical metrics such as the correlation coefficient (r), the root mean squared error (*RMSE*), the root mean squared percentage error (*RMSPE*), squared difference between standard deviations (SDSD, Kobayashi and Salam, 2000) and lack of correlation weighted accuracy of the model along with the requirements for by the standard deviations (LCS, Kobayashi and Salam, 2000). The comparison to both the $RH_{98}$, as well as the ALS and VHR-HM measurements will be conducted at the resolution of 10m, which is the target resolution of our models. This should not pose any issues since the ALS and VHR-HM are at a much higher resolution and were downsampled to 10m. However, GEDI measurements are acquired over 25m wide footprints. As such, they contain height information from ~2×2 pixels from our proposed 10m height maps. Nonetheless, photons from GEDI pulses are mostly packed within the center of the footprint, and contribute the most on the measured height profile, with weaker contribution alongside the borders of the footprint. Moreover, over homogenous areas, the measured maximum height at any given ~2×2 pixels within the footprint should be similar. Therefore, in this study we assume that a pixel size of 10×10 m should be representative of the measured maximum height by GEDI. *r*, RMSE, RMSPE, SDSD, and LCS are defined in Appendix H.

Next, we will evaluate the proposed models based on their output image sharpness index ($SI_o$) in comparison to the image sharpness index from the S2 images ($SI_{S2}$). The ratio of $\frac{SI_o}{SI_{S2}}$, which we refer to as the Ground Sampling Index (GSI) will provide an indicator of the real resolution of the different products. In fact, while all the models were designed to produce height maps with a pixel size of 10m, the real resolution might be lower due to several reasons, such as the inherit qualities of the proposed models' building blocks, or the models' attempt to remove noise and artifacts. The sharpness index of the outputs and S2 images are based on the perceptual sharpness metrics (SM) proposed by Crete et al., 2007. In our formulation, GSI is calculated over 256×256 patches, $SI_o$ is the SM calculated from a given model's output while $SI_{S2}$ is the average SM value from the 10 m S2 bands over the same scene. GSI thus ranges from 1 to +∞, with GSI = 1 representing a scene with a GSD identical to S2. Since GSI does not scale linearly with the resolution, we calculated several GSI values for several smoothed S2 images at different resolutions ranging from 12.5 m down to 40 m. GSI of images at full resolution (i.e., 10m) as well as down-sampled S2 images can be found in Table 2.

| **Resolution (m)** | 10 | 12.5 | 15 | 17.5 | 20 | 22.5 | 25 | 27.5 | 30 | 32.5 | 35 | 37.5 | 40 |
|---|---|---|---|---|---|---|---|---|---|---|---|---|---|
| **GSI** | 1 | 1.03 | 1.08 | 1.14 | 1.21 | 1.29 | 1.37 | 1.46 | 1.56 | 1.68 | 1.77 | 1.88 | 2.00 |

Table 2. GSI at different resolutions computed by averaging the GSI over 5000 256×256 S2 image patches across different landcover types. S2 image patches at resolutions lower than 10m were obtained by Lanczos interpolation.

Finally, the resulting height maps from our proposed models are compared to two global height products, the product of Lang et al., 2022 (using S2 as inputs; hereafter referred to as Lang), at 10m resolution, and that of Potapov et al., 2021 (using Landsat as inputs; hereafter referred to as Potapov) at 30m resolution. The comparison is based on their accuracy in comparison to the ALS dataset, their GSI, as well as the quality of the output maps.

*4.4. Results*

*4.4.1. Model performance in comparison to GEDI*

We evaluate the performance of the four deep learning models (2MOU, 2MDU, α-2MDU, and Hy-TeC) in comparison to the set-aside GEDI dataset. The GEDI acquisitions and the corresponding 256×256 optical and radar imagery, randomly selected from the larger 768×768 tiles, were chosen to represent a large variety of biomes (e.g., savannas, deciduous and evergreen forests).

The performance of the first tested deep learning model 2MOU (Fig. 8), which is optimized using a Huber loss, shows that the model exhibits clear signs of saturation starting at ~30m (Fig. 8, box plots). Indeed, the bias (difference between $RH_{98}$ and 2MOU $RH_{98}$) increased from 1.6 m for $RH_{98} \in [20m, 30m]$ to 14.7 m for $RH_{98} \in [45m, 50m]$. The results also show that the model, while able to distinguish between forest/non-forest areas, failed to capture the variability of the canopy heights. In fact, for tall canopies ($RH_{98} > 20m$), the model appears to only produce estimates centered around the mean height (~30m). As such, to reduce the overall errors on the tall trees ($RH_{98} > 35m$) the model over-estimated the shorter ones ($RH_{98} \in [15m, 25m]$). This is observed by the RMSPE that decreases from 24.1% for $RH_{98} \in [15m, 20m]$ to 19.8% for $RH_{98} \in [30m, 35m]$ before increasing again for $RH_{98} > 35m$. For areas with bare soil/low vegetation and short trees ($RH_{98} < 15m$) the model appears to accurately estimate GEDI's $RH_{98}$ for the majority of acquisitions. Nonetheless, a significant number of estimates show large discrepancies ($|2MOU_{RH98} - RH_{98}| > 15m$).

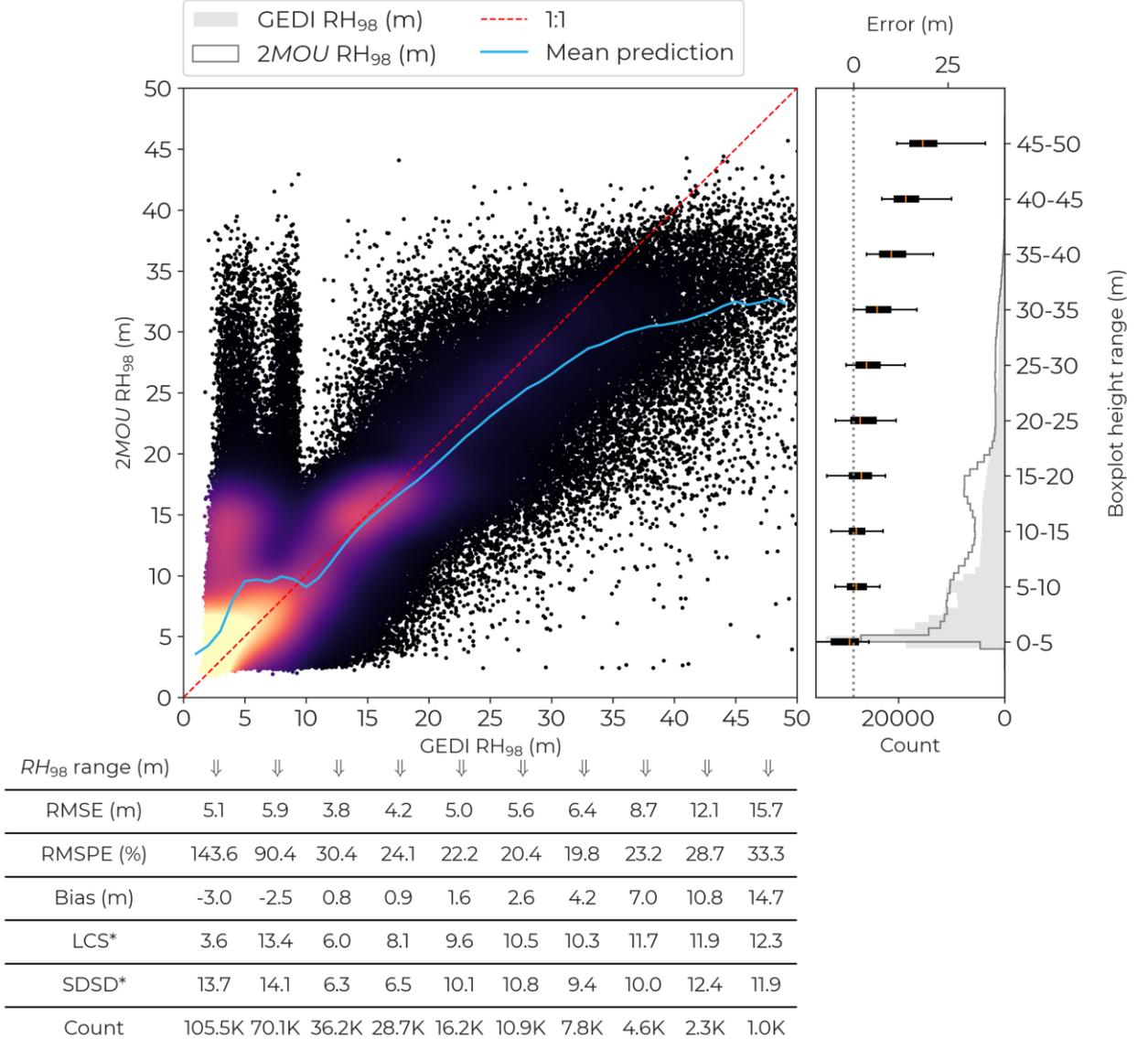

Figure 8. Performance of the 2MOU model in comparison to GEDI's $RH_{98}$. The summary-statistics table is calculated by ranges of $RH_{98}$ (bounded by the x-axis tick positions).

For these estimates, the error sources could mostly be classified into three categories: Category I (cat. I, Fig. 9) represent cases of instrument errors where GEDI failed to produce high quality waveforms but were good enough to pass our filters. The poor quality of such waveforms is either due to insufficient penetration (e.g., cat. I.1, Fig. 9), or due to unfavorable acquisition conditions, and as a consequence severely underestimated the canopy height. For example, cat. I.2 and I.3 (Fig. 9) shows two instances where GEDI estimated the height of tall trees as being lower than 6m, while the model accurately produced high values. Category II errors (cat. II, Fig. 9), represents the case where GEDI acquired high quality waveforms, but with large errors on the geolocation of the footprint. For example, in Fig. 9 cat. II.1 and cat.II.3 show two waveforms with a single peak (i.e., bare soil/low vegetation) while the footprint location shows trees. In contrast, Fig. 9 cat.II.2 represents a waveform with two distinct peaks (indicating a tree), while the footprint location is over bare soil. Category II errors concerns mostly footprints acquired over Savannas, where the tree cover is sparse. Finally, Category III errors (cat. III, Fig. 9) represent the case where GEDI accurately measured the vertical structures within the footprint with low errors on geolocation, however, given the small size of the canopies (characterized by a low vegetation peak in the waveform) and the low resolution of S2 and S1 (10m), the model was unable to accurately predict the height of these trees.

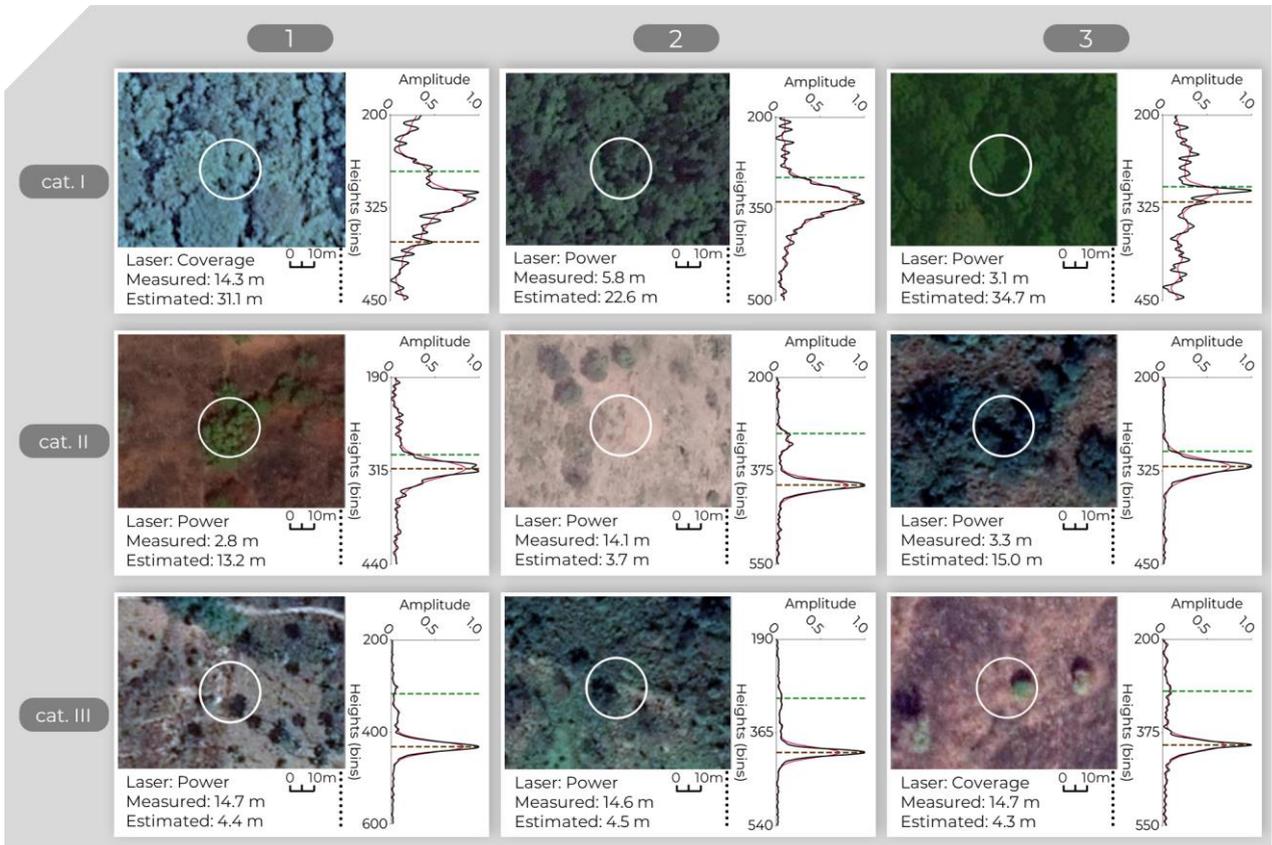

Figure 9. Example of several GEDI waveforms (black line) across different landscapes where the 2MOU model produced height estimates with large differences in comparison to GEDI's $RH_{98}$. Footprint location is shown in white; detected ground and canopy top returns derived from the smoothed waveform (red) are respectively shown in brown and green.

### 4.4.2. 2MDU performance

The 2MDU model employs the same encoder and decoder as the 2MOU model, however it has a modified output head that constrains the estimated outputs to ten overlapping height ranges before regressing the height estimate. This translates to a high increase of accuracy across all height ranges in comparison to the 2MDU model (Fig. 10). Indeed, performance results of the 2MDU shows that the model no longer saturates over tall trees ($RH_{98} > 30$m, Fig. 10 boxplots), and overall has low RMSE that ranges between 3.0 m ($RH_{98} < 5$m) to 5.7 m ($RH_{98} > 45$ m). Moreover, results also show that both the LCS, and SDSD are mostly similar across all height ranges, indicating that the model was able to capture both the magnitude and pattern of the fluctuations to the same degree for each height range. In terms of bias, the model appears to overestimate $RH_{98}$ values smaller than 10m and underestimate those higher than 10m. However, for $RH_{98} > 10$ m, the bias is mostly constant and ranges between 1.7 and 2.5 m (Fig. 10).

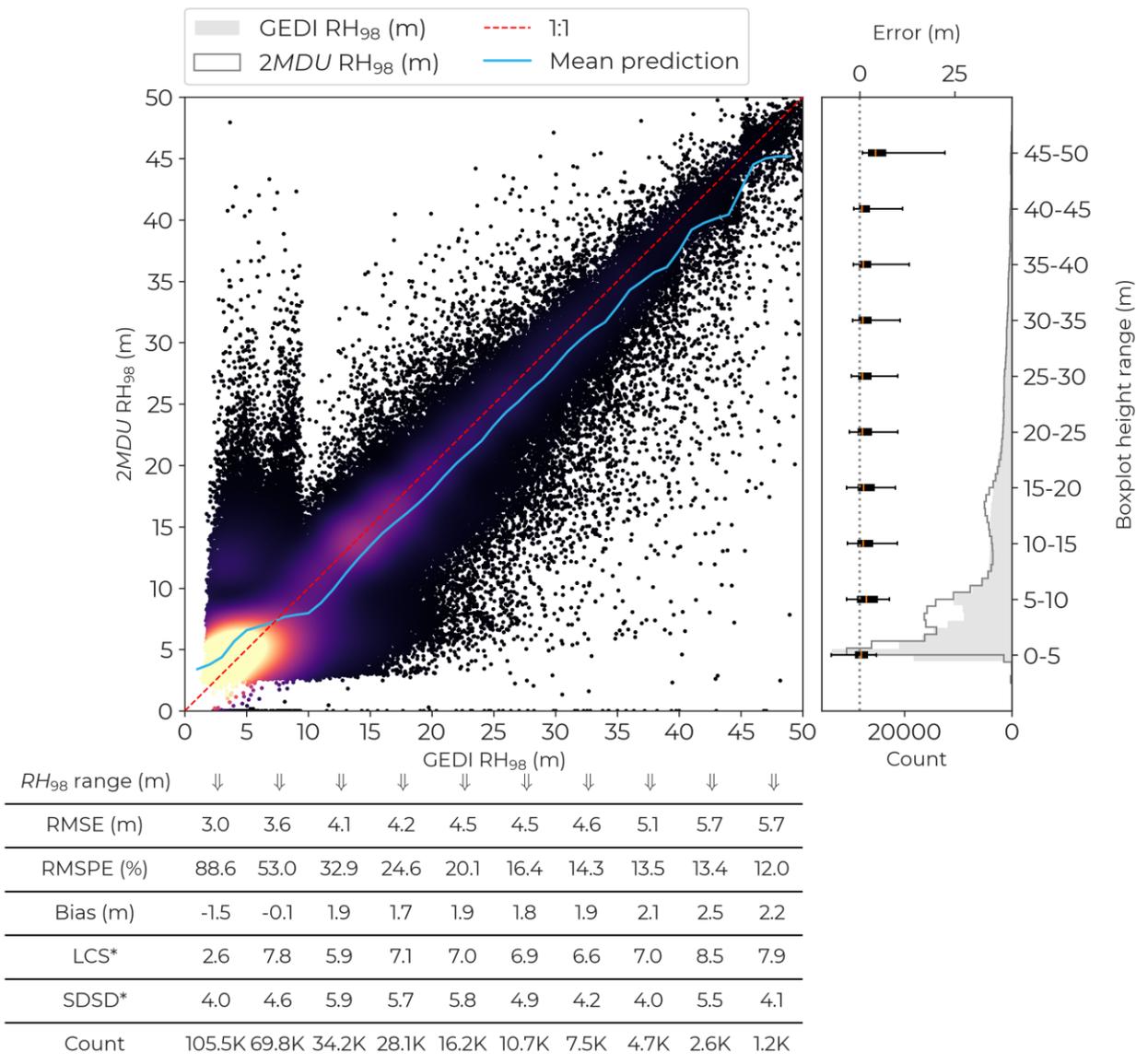

Figure 10. Performance of the 2MDU model in comparison to GEDI's $RH_{98}$. The summary-statistics table is calculated by ranges of $RH_{98}$ (bounded by the x-axis tick positions).

Nonetheless, as this model learned to better fit the GEDI acquisitions, it also learned to fit well the outliers. This is especially true for the category I errors (Fig. 9, cat.I) where the RMSE decreased from 5.1 and 5.9m for respectively $RH_{98} \in [0m, 5m]$ and $RH_{98} \in [5m, 10m]$ (Fig. 8) to respectively 3.0 and 3.6m (Fig. 10). As a consequence, the mapping capabilities of the 2MDU model degraded as it started predicting the height of some forest patches as being bare soil/low vegetation (Fig. I.25).

*4.4.3. α-2MDU performance*

The α-2MDU model shares the exact same architecture as the 2MDU model, however, the main difference between the two models lies in the cost function associated with the regression part. The 2MDU uses the Huber loss, while the α-2MDU employs the adaptive loss which reduces the sensitivity of the outliers as training progresses (Fig. J.26). As a consequence, the accuracy of the α-2MDU model in comparison to the set-aside GEDI dataset is lower relatively to the 2MDU model across all height ranges (higher RMSE and RMSPE in general, refer to Figs. 10 and 11). However, the difference in accuracy (ΔRMSPE) between the two models is <11% for $RH_{98} \geq 10m$. Moreover, the bias is reduced with the α-2MDU model for $RH_{98} \in [10m, 40m]$ in comparison to the 2MDU model, after which, the model appears to underestimate taller trees with a bias of respectively 3.4m (RMSE = 7.0m) and 5.7m (RMSE = 9.1m) for $RH_{98} \in [40m, 45m]$ and $RH_{98} \in [45m, 50m]$.

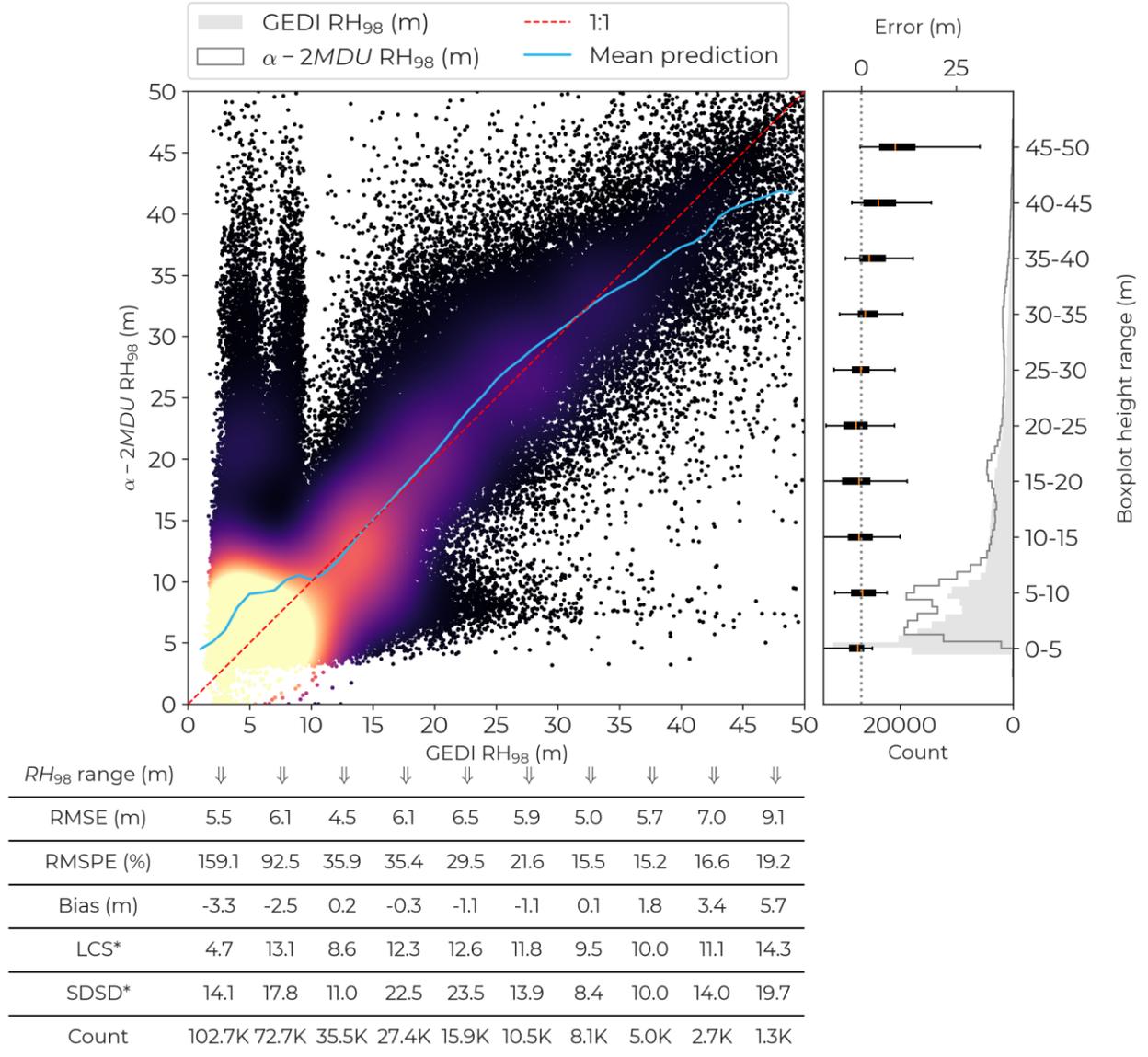

Figure 11. Performance of the α-2MDU model in comparison to GEDI's $RH_{98}$. The summary-statistics table is calculated by ranges of $RH_{98}$ (bounded by the x-axis tick positions).

Finally, as the α-2MDU model is less sensitive to outliers, especially cat. I errors (Fig. 9), the model no longer shows artifacts over dense tropical forests (Fig. 12).

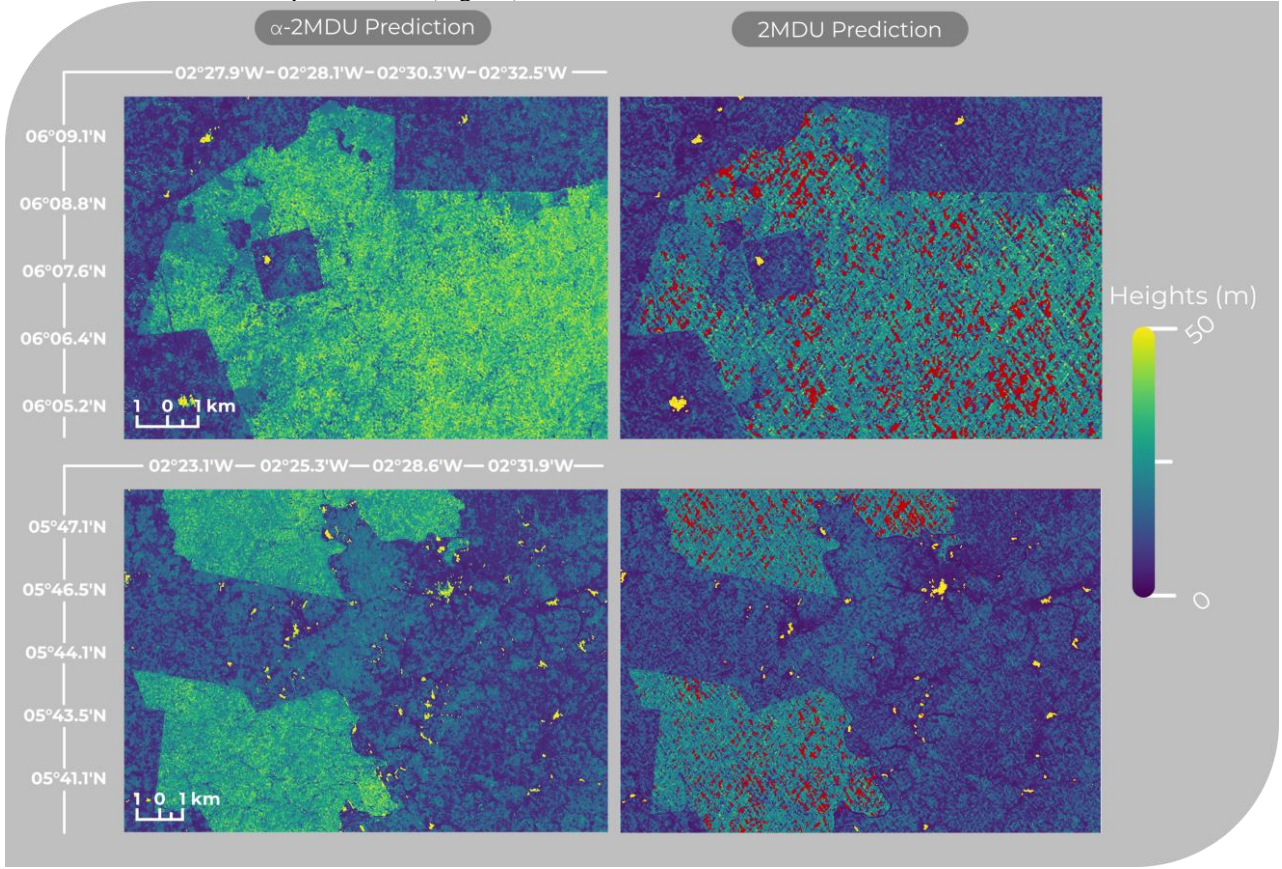

Figure 12. Comparison between the predictions of the α-2MDU and 2MDU models over two areas with tall trees. The red polygons represent forested areas where the 2MDU model estimated the heights of tall tress as being less than 2m.

### 4.4.4. Hy-TeC Performance

The Hy-TeC model which utilizes the same output head and loss functions for its main output (10m) as the α-2MDU model, shows slightly improved accuracy than α-2MDU in comparison to the set-aside GEDI dataset (Fig. 13). For $RH_{98} < 10m$, Hy-TeC seems to be less sensitive to cat. I outliers (Fig. 9) in comparison to α-2MDU (RMSE higher by 0.2m for the Hy-TeC model, Figs. 13 vs. 11). For $RH_{98} \in [10m, \ 30m]$ the RMSE on these estimates by Hy-TeC is lower than those provided by α-2MDU by a slight margin (less than 0.4m), while for $RH_{98} \in [30m, \ 40m]$, the α-2MDU estimates appear to be marginally more accurate than those provided by Hy-TeC by up to 0.4m. For $RH_{98} > 40$ m, Hy-TeC shows better estimation capabilities, with lower RMSE (7.2m vs. 7.8m) and lower bias (2.8m vs. 4.1m). In terms of LCS and SDSD, Hy-TeC has mostly lower values than α-2MDU expect for $RH_{98} \in [30m, 40m]$ (Fig. 13 vs. Fig. 11), however the differences are minimal between both models across all height ranges.

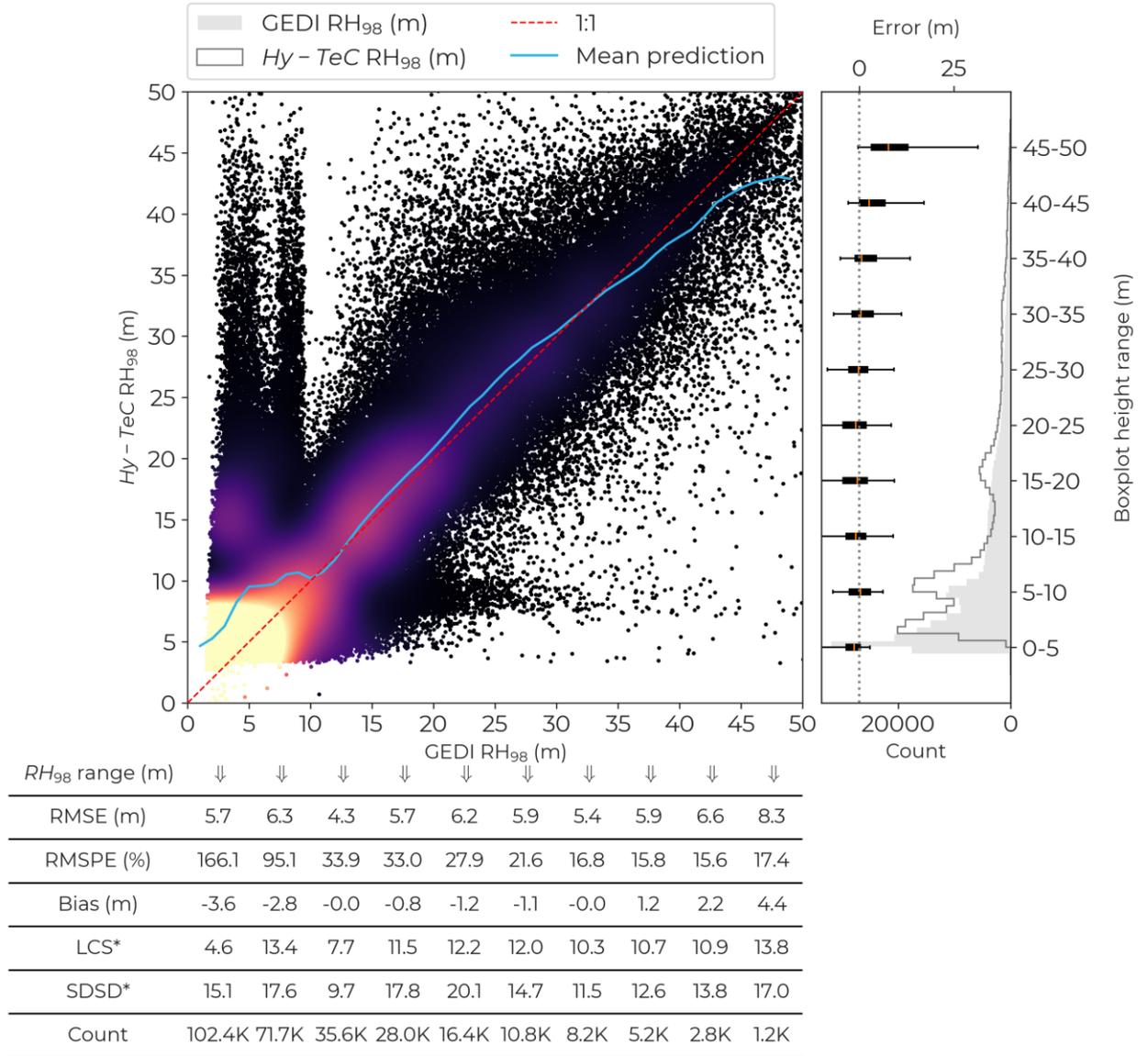

Figure 13. Performance of the Hy-TeC model in comparison to GEDI's $RH_{98}$. The summary-statistics table is calculated by ranges of $RH_{98}$ (bounded by the x-axis tick positions).

## 4.5. Comparison to ALS and VHR-HM

The analysis made in the previous sections showed that two models presented better performances than the others: α-2MDU and Hy-TeC. These two models are selected in this evaluation section.

The performance of our two best models, namely the α-2MDU and Hy-TeC, are evaluated against an independent ALS campaign (Stereńczak et al., 2020) and the VHR-HM map. Both datasets were upscaled to a resolution of 10m to coincide with the theoretical resolution of our output maps. For comparison purposes, we also evaluated three other datasets to the ALS and VHR-HM datasets. The first dataset represents the GEDI acquisitions over the ALS area (resp. VHR-HM area) where we scaled the ALS (resp. VHR-HM) resolution to 25m to coincide with the 25m wide footprints. The second and third datasets are the 10m global height map from Lang et al., 2022 and the 30m global height map from Potapov et al., 2021, respectively (ALS and CHR-HM upscaled to 30m for Potapov).

The upscaling of the ALS (resp. VHR-HM) from 1m (resp. 0.7 m) to another resolution is as follows: for a given pixel from an evaluated map, we selected all the ALS pixels (resp. VHR-HM) within that pixel, calculated the cumulative density function (CDF) at 0.1m intervals, and selected the position at 97% of the CDF, which correlated best with GEDI's $RH_{98}$.

The results presented in Fig. 14.a show that compared to the ALS dataset, the GEDI acquisitions, which were used as calibration data for the four models, underestimated the ALS measurements by 3m (bias), with an RMSE of 8.93m and a correlation coefficient (r) of 0.18. Separating the GEDI acquisitions into Coverage (Cov) and Full Power (FP) acquisitions, the results show that acquisitions from the FP lasers are generally more accurate, with an RMSE of 7.9m (r=0.30), while the RMSE for the Cov acquisitions is 9.6m and r = 0.17. Although both the FP and Cov acquisitions underestimated the ALS measurements, the FP showed a lower bias (2.2m) compared to the Cov (bias= 5m).

The results of the modeled heights from the four products show that Potapov's 30m map is the least accurate one with an RMSE of 10.0m, a correlation coefficient (r) of 0.11, and a bias of 6.1m (Fig. 14.b). The 10m height map of Lang shows better accuracy in comparison to Potapov, with a lower RMSE (8.6m), higher correlation coefficient (r=0.18), and lower bias (4.9m) (Fig. 14.c). However, the results also show that both height products are unable to produce height estimates above 40m.

Finally, the results show that our two proposed best models, α-2MDU and Hy-TeC (respectively Fig. 14.d and Fig. 14.e), are the most accurate ones, with Hy-TeC showing slightly better performance with a respective RMSE of 8.3m (r=0.22) and 7.9m (r=0.25) for the α-2MDU and Hy-TeC models. However, the bias of the height estimates from the Hy-TeC model was marginally higher (0.15m) than that of the α-2MDU model. Moreover, unlike Potapov's and Lang's products, the α-2MDU and Hy-TeC models do not appear to saturate for tall trees as both were able to produce canopy height estimates higher than 45m.

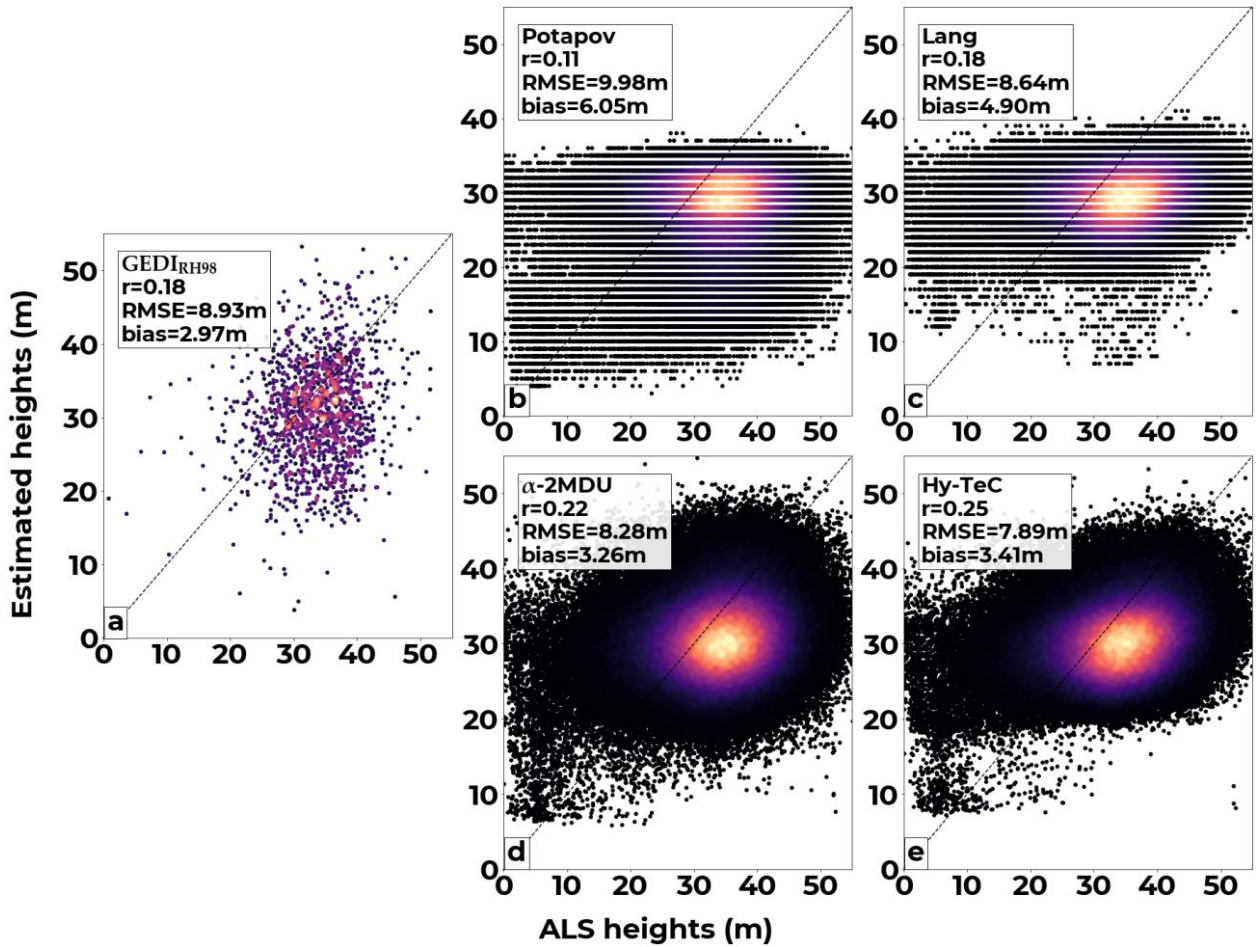

Figure 14. Comparison of estimated heights from (a) GEDI, (b) Potapov, (c) Lang, (d) α-2MDU, and (e) Hy-TeC *vs* the reference ALS data set.

In the evaluation *vs* the reference VHR-HM dataset, over sparsely vegetated areas, GEDI's $RH_{98}$ measurements show better accuracy in comparison to the reference dataset than over densely vegetated areas with a correlation coefficient of 0.48, a RMSE of 3.9m and a bias (VHR-HM $-RH_{98}$) of -1.3m (Fig. 15.a). Regarding model performance (Figs. 15.b, 15.c, 15.d and 15.e), Hy-TeC shows the highest accuracy on the estimation of the heights in a sparsely vegetated context with a correlation (r) to the reference dataset of 0.56, a RMSE of 3.1m and bias of -1.7m (Fig. 15.e). α-2MDU's performance (Fig. 15.d), while marginally better than Hy-TeC in terms of correlation to the reference dataset, the accuracy on the height estimates with α-2MDU was significantly lower than Hy-TeC, with a RMSE of 4.3m and a bias of -3.4m. Regarding the global products, the Potapov product at 30m (Fig. 15.b) showed almost no bias to VHR-HM, however, their product has low correlation (r=0.29) and high RMSE (6.8m). Finally, for the Lang product (Fig. 15.c), while the height estimates showed a correlation to the reference dataset (r=0.49) that is comparable to what is obtained with GEDI (r=0.49), their product has a very high bias (-8.1 m), which in turn decreased significantly the accuracy of their estimates (RMSE=8.6m).

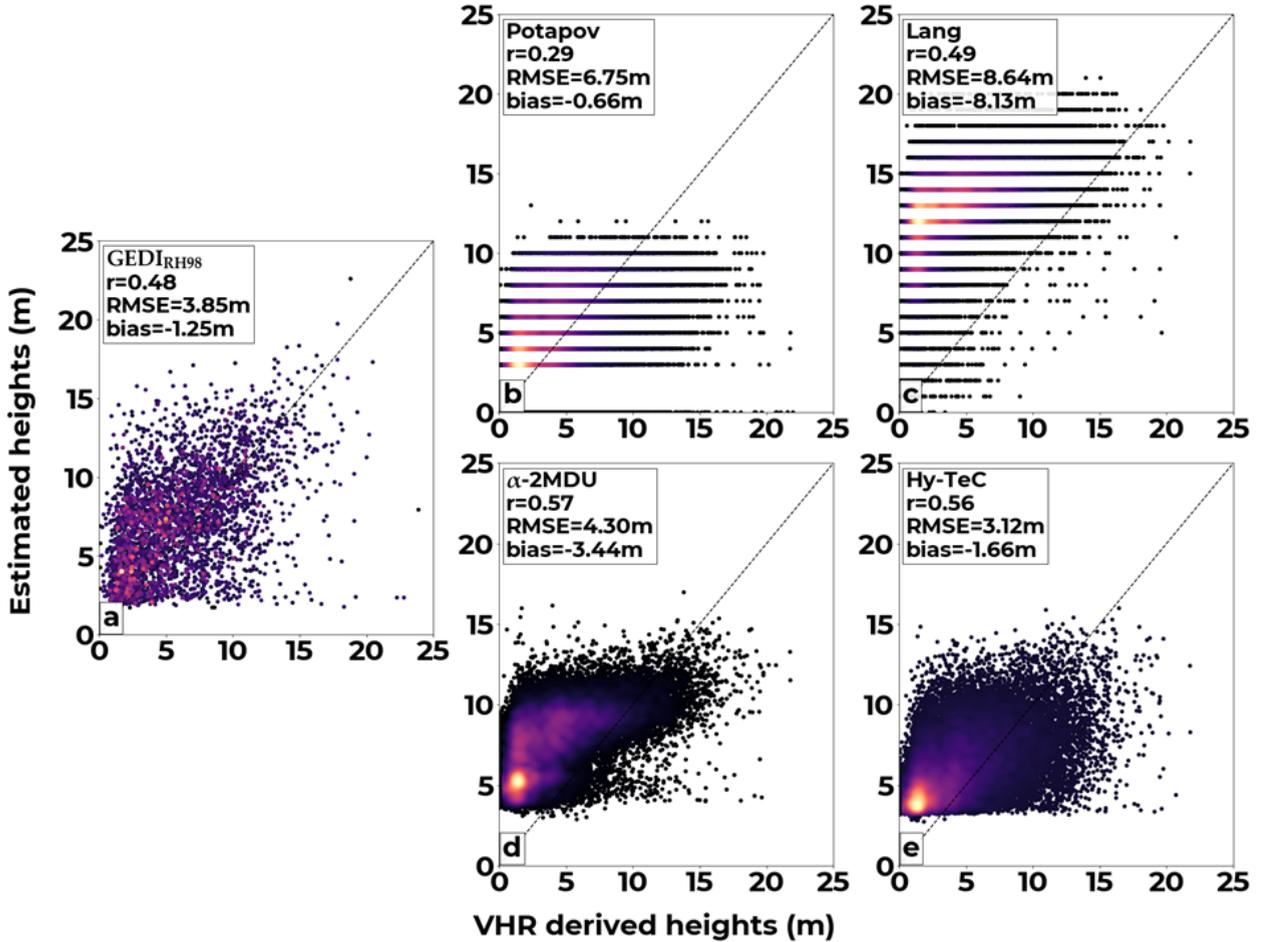

Figure 15. Comparison of estimated heights from (a) GEDI, (b) Potapov, (c) Lang, (d) α-2MDU, and (e) Hy-TeC *vs* the reference $VHR$-HM data set.

## 4.6. Height maps quality assessment

In this section, we present a comprehensive evaluation of four different canopy height map products, including our two proposed best models (α-2MDU and Hy-TeC) and two global products (Potapov and Lang). The evaluation is divided into two parts: first, we assess the ground sampling index (GSI) of each product to evaluate their real output resolution, and second, we conduct a qualitative assessment of the maps to assess their overall quality.

Regarding the ground sampling index (GSI, section 4.3) of the different products, results presented in Fig. K.27 shows that from the three models with an output pixel size of 10m (Hy-TeC, a-2MDU and Lang) , Hy-TeC has the lowest GSI, with 80% of patches having a GSI lower than 1.08 which is comparable to the GSI of S2 patches with a resolution between 10 and 15m (Table 2). For the remaining 20% patches, 10% of patches have a GSI between 1.08 and 1.14 (15 and 17.5m, Table 2), while the remaining 10% have a GSI between 1.14 and 1.46 (17.5 and 27.5m, Table 2). The α-2MDU model has a higher overall GSI in comparison to Hy-TeC, with 58% of patches showing a GSI between 1 and 1.08 (between 10 and 15m, Table 2), 32% of patches have a GSI between 1.08 and 1.21 (15 and 20m, Table 2), while the remaining 10% patches have a GSI between 1.21 and 1.46 (between 20 and 27.5m, Table 2). For the Lang model, 25% of the patches have a GSI between 1 and 1.08 (between 10 and 15m, Table 2), 54% between 1.08 and 1.56 (between 15 and 30m, Table 2), and the remaining patches have a GSI between 1.56 and 2.0 (between 30 and 40m, Table 2). For the 30m height maps from the study of Potapov et al., 2021, the GSI of the tested patches appear to be close to the reported resolution for the majority of the cases (80%, GSI between 1.46 and 1.68, Table 2), while the remaining patches showed a GSI between 1.68 (32.5 m, Table 2) and 2.0 (40 m, Table 2).

The higher resolution of our two proposed models can be clearly seen in the resulting height maps. For example, over dense forested areas (Fig. 16), both Hy-TeC and α-2MDU models clearly show higher ground sampling distances and this is apparent from the 3D-renders with denser predictions over the forest patches (Fig. 16, 3D-Render). Moreover, in comparison to the distribution of heights from the GEDI acquisitions (Fig. 16, Height distributions), both, Hy-TeC and α-2MDU show a higher capacity to predict both low (bare soil/low vegetation) and high values (> 50 m). In contrast, the Potapov model appears to saturate above 35 m, while the Lang model shows height estimates ranging between 10 m (no low values) and 45 m.

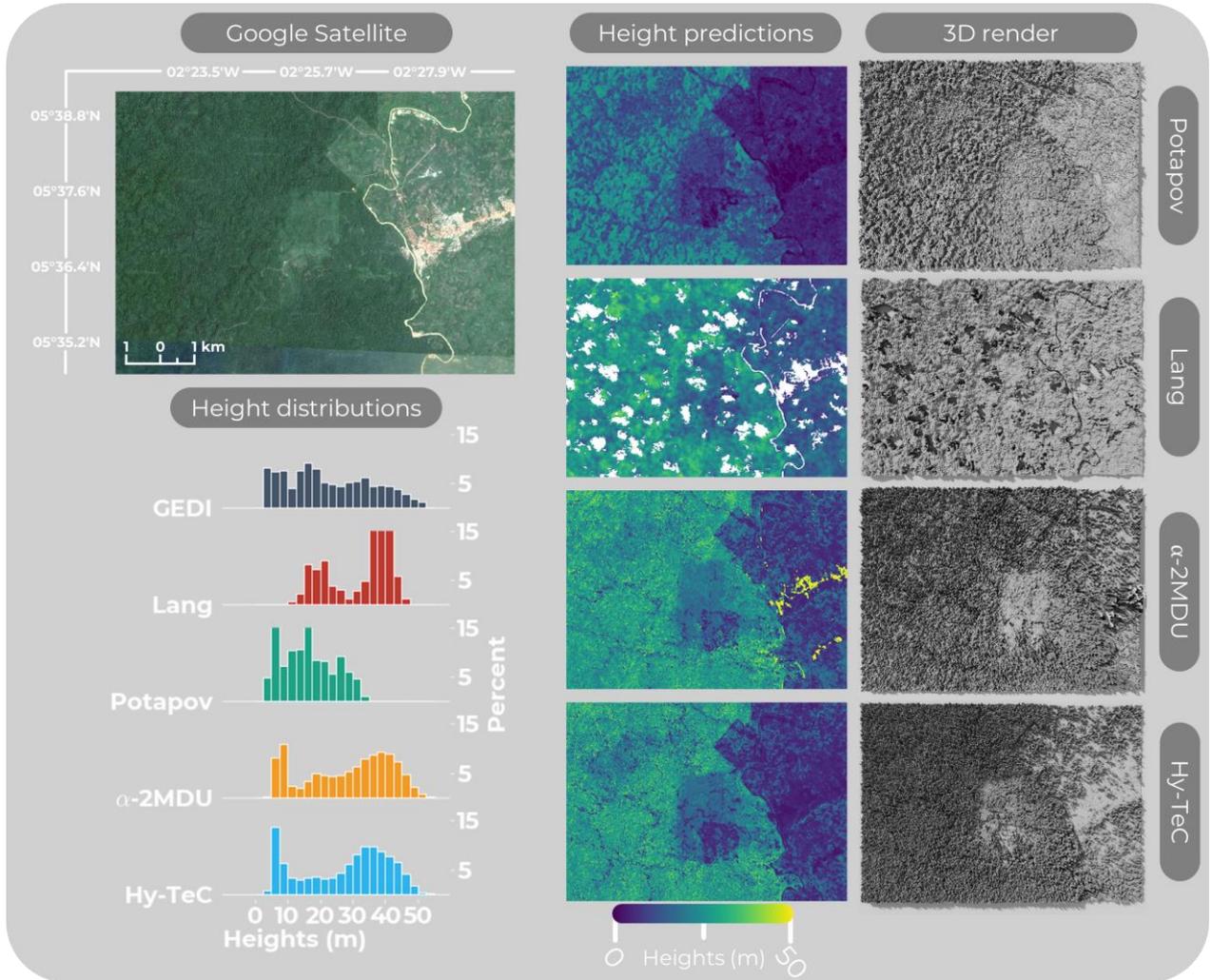

Figure 16. Land cover height predictions (Height Predictions) from GEDI estimates and from the four compared models (Potapov, Lang, α-2MDU and Hy-TeC) over an area with predominant forest areas. 3D Render, is a 3D visualization of the land cover estimates. The Height distributions represent the distribution of the land cover heights from each of the four products as well as from the available GEDI acquisitions.

Nonetheless despite the overall agreement between the α-2MDU and Hy-Tec maps, there appears to be some differences between the two models. First α-2MDU shows slightly higher estimates for tall trees and also for short heights (< 10m) than Hy-TeC. The higher ground sampling distance is also apparent, both in the 3D-renders (Fig. 16), as well as the fine trails within the forest patch (Fig. 16, Height predictions) that are better captured by Hy-TeC than α-2MDU.

Over lightly forested areas, Fig. 17 shows the improved height estimation capabilities of our two proposed models in comparison to current available products. Overall, Potapov's model shows lower estimation of canopy heights than the three other models. Nonetheless, both Potapov and Lang show estimates which are far smoother than ours. Moreover, both Potapov and Lang also seem to miss most of the small vegetation.

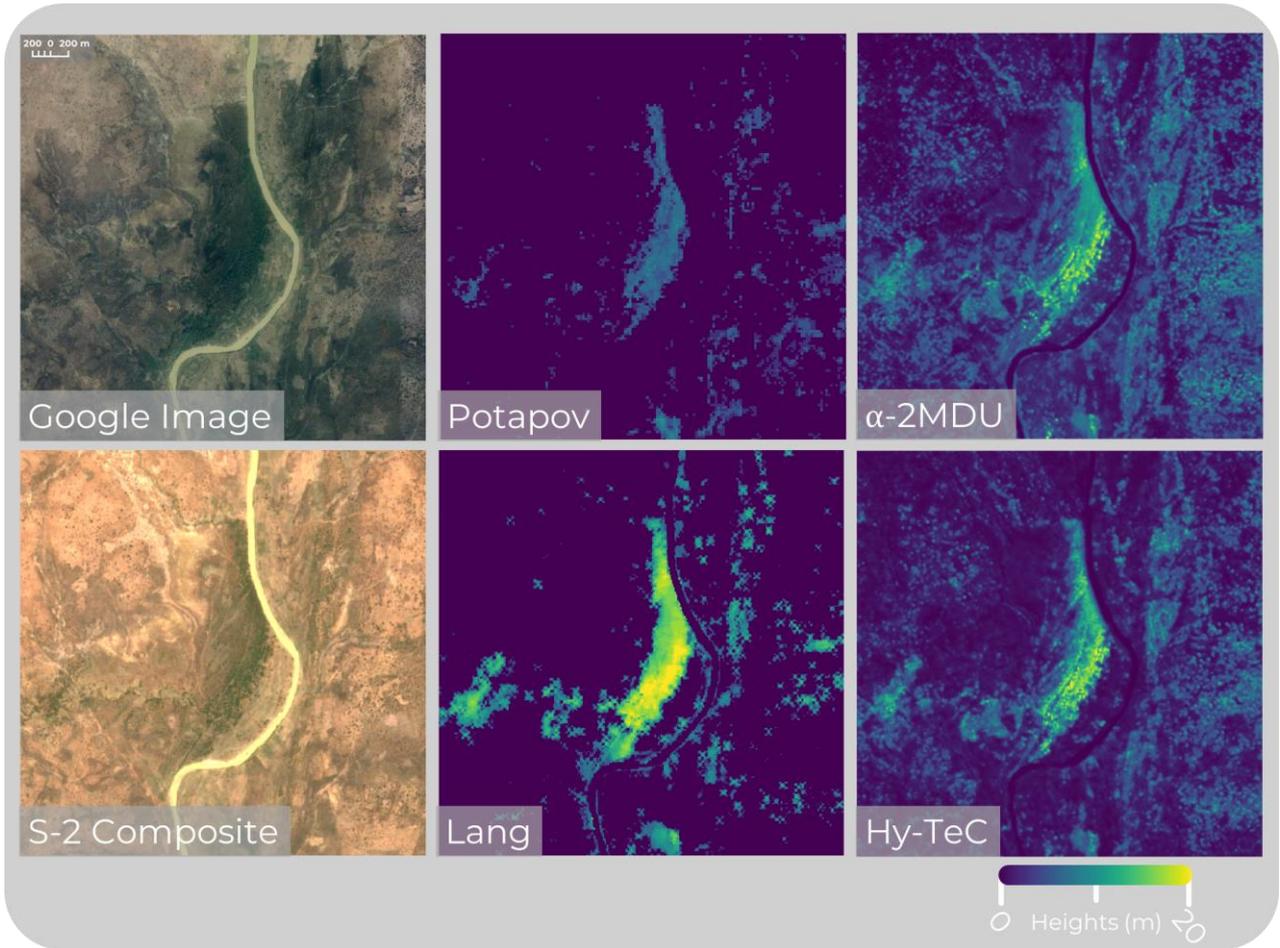

Figure 17. Illustration of the land cover height predictions from the four compared models over an area with sparse tree cover.

The differences between our proposed models and existing products are more apparent in more open areas. Indeed, while α-2MDU and Hy-TeC appear to estimate the heights of most trees in Fig. 18, the models of Potapov and Lang only estimated the heights of the largest ones.

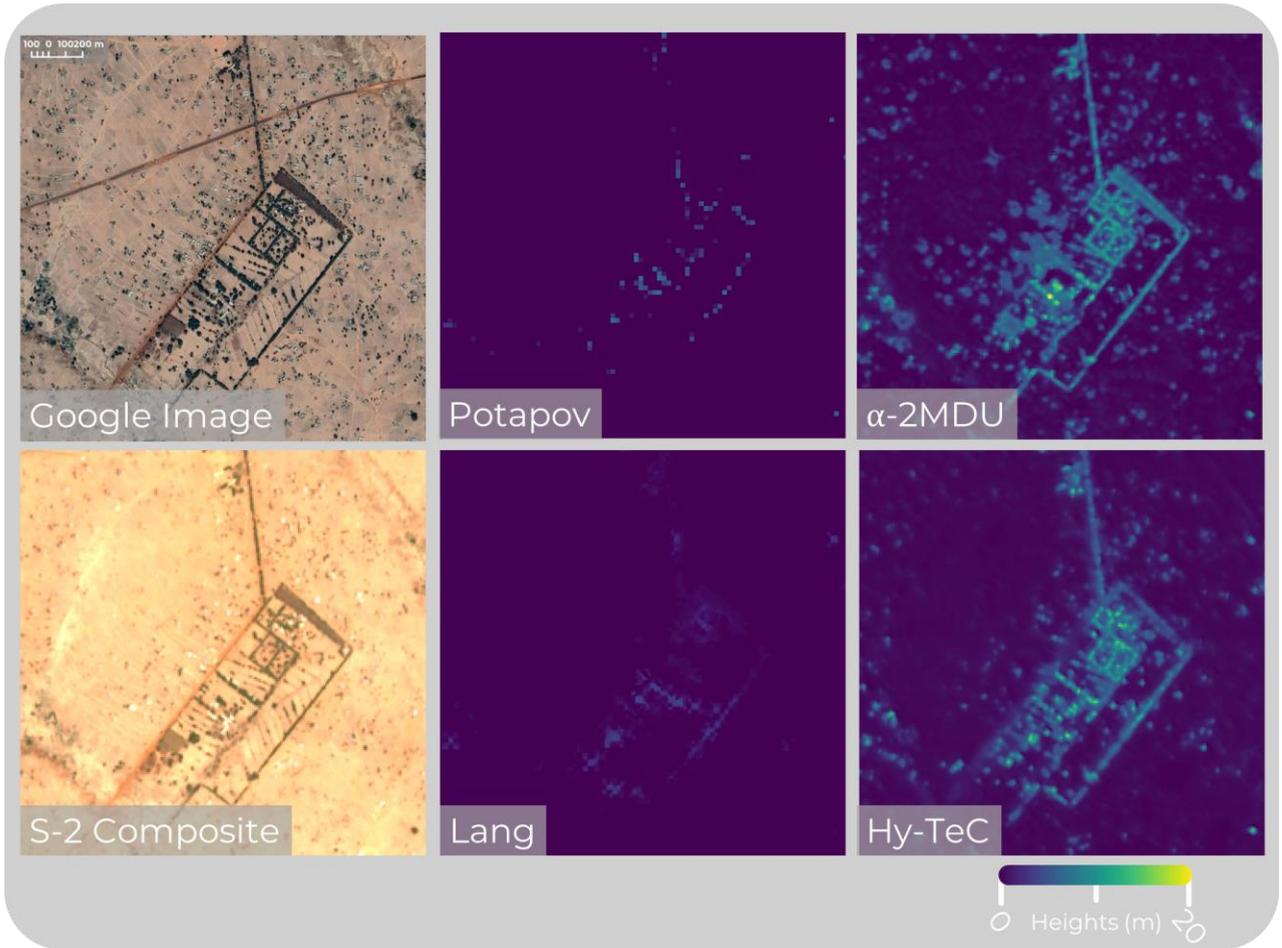

Figure 18. Illustration of the land cover height predictions from the four compared models over planted trees.

Over urbanized areas, Hy-Tec seems to perform the best (Fig. 19). Indeed, Hy-TeC accurately removed all built-areas from its height predictions and was still capable of predicting tree heights from isolated trees and forest patches. The model of Potapov also seems to perform well over urban areas; however, it produces heights at coarser resolutions. Finally, for α-2MDU, while the predictions appear to resemble those of Hy-TeC over forested areas within the urban scene, it shows two distinct differences in comparison to Hy-TeC. First, α-2MDU appears to provide in general higher height estimates for forested areas, and over built-up it predicts very high values with no patterns, and thus it will require a post-processing step to mask these predictions. This also seems to be the case for the Lang model (Fig. 19), as the white spots indicate that urban areas have been masked in post-process.

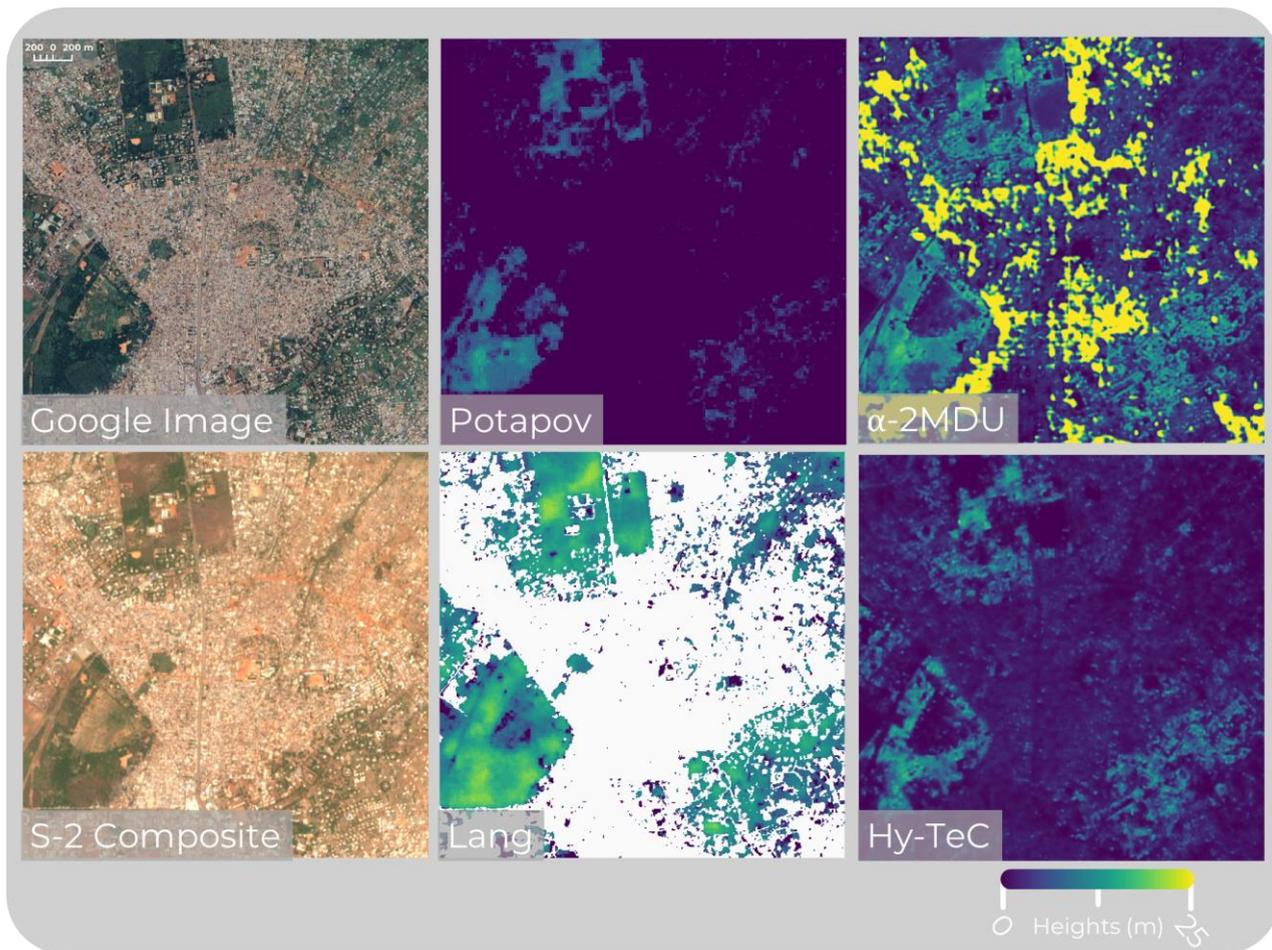

Figure 19. Illustration of the land cover height predictions from the four compared models over urban areas.

## 5. Discussion and conclusions

In this study we presented and evaluated two deep learning models, α-2MDU, a model based on the U-Net architecture, and Hy-TeC, a hybrid model that employs a transformer encoder and a ConvNet decoder. We found that these models which use as inputs S2 and/or S1 acquired imagery and calibrated with GEDI acquisitions provide significant improvements over current state-of-the-art approaches both quantitatively and qualitatively. Specifically, evaluated against the set aside GEDI dataset, ALS dataset, and the VHR-HM dataset, our models have shown significantly reduced saturation on the height estimates of tall trees, and due to their higher output resolutions, our approaches allow better detection and height estimation of trees in sparsely vegetated areas. The models are also all-terrain models and make no assumptions when estimating the heights of the underlying land cover, reducing the need for additional data sources and further pre- or post-processing. However, it is important to acknowledge that each of the two models have its own uncertainties and limitations that are influenced by various factors, including the pre-processing of the S1 and S2 images, the quality of GEDI data, model architecture, and the loss functions. In this section, we will discuss the impact of these factors on model uncertainty and explore possible approaches to further reduce it, ultimately improving the accuracy of the predictions.

*5.1. S1 and S2 image pre-processing*

Our main inputs for the deep learning models were S1 and S2 image mosaics acquired over a 15 months period. Mosaicking presents advantages as well as drawbacks. In terms of advantages, especially for optical imagery, a mosaic for a given time period allows the reduction of several problems including cloud contamination, surface directional reflectance, and view and illumination geometry (Flood, 2013). This proved beneficial in our case, especially over the south of Ghana, a cloudy region where our models were able to produce continuous and artifact-free canopy height estimates over the dense tropical forests. In the literature, several techniques were proposed to create a mosaic from a times series, the maximum NDVI composite (MNC; Holben, 1986; Huete et al., 2002; Roy et al., 2010), medoid composites (MC), a multi-dimensional median technique (Flood, 2013; Kennedy et al., 2018), or simply the median (Liang et al., 2023; Schwartz et al., 2022). Here we decided to use the median instead of MNC or MC for several reasons. First, one of the major benefits of the MNC technique is its capacity to reduce cloud contamination and cloud shadows. However, with the advancement of cloud and cloud shadow detection techniques (Goodwin et al., 2013), and the fact that MCN will produce mosaics biased towards high NDVI values, potentially reducing the contrast between forests and grasslands, the MCN was not adopted. Second, the MC technique, in contrast to the median, will select a single date for each pixel as the representative of the time-series, as such, will better preserve the relationship between the bands, whereas the median method will calculate a median value for each band independently. On the other hand, the MC is more computationally expensive than the median, and while band correlation might be higher with MC, we do not believe this should have a significant impact on the height estimates, as our models do not rely on calculated vegetation indices, but rather on raw reflectance and backscatter values used as inputs. Regarding the S1 data, while mosaicking could help reduce speckle noise, in the general case, it is not necessary as SAR instruments have an all-weather operability. However, due to terrain changes that might have occurred during the 15 months period, we also created a median mosaic for S1 data to stay consistent with S2 data.

In terms of drawbacks, a mosaic over a time period, especially over long time periods can potentially mask large events within a landscape, which might not be desirable depending on the target application. For example, for forest monitoring applications, events such as fires or logging, especially those taking place at the end of the considered time period for the mosaic, will not be detected until later. For agricultural applications, where planted crops are usually rotated each season, a mosaic of more than a season will mix information from different types of planted species, which is not desirable. As such, a compromise between image quality and temporal fidelity should be made.

*5.2. GEDI as calibration data*

Currently GEDI is one of only two operating spaceborne LiDAR instruments with quasi-global coverage and will remain an essential source of information on forest vertical structures in the near future. However, data acquired by GEDI could potentially be contaminated by large sources of errors, impacting the accuracy of the produced height maps. First, GEDI is a full waveform instrument, and with its pulse width of 14 ns (~ 4m; Dubayah et al., 2020), GEDI is less sensitive to vertical structures that are shorter than 4m. Therefore, by relying on the $RH_{98}$ metric as calibration data, our proposed models are overestimating heights over bare soils, water bodies, and agricultural areas. Indeed, the lowest estimated height over Ghana was around 2.5m over water surfaces and around 3m over bare soils/low vegetated areas. These uncertainties could be reduced with different approaches. Di Tommaso et al., 2021 proposed an approach to classify the GEDI shots as being acquired over tall or short crops, which could further be developed to include water bodies and bare soils. A pre-processing step, before model calibration, would then be to assign a lower height value (e.g., a lower RH metric) for the GEDI shots belonging to the low height classes (e.g., short vegetation, water bodies, bare soil). Another approach would be to replace GEDI acquisitions for short vegetations (e.g., less than 25m) by ICESat-2 acquisitions. ICESat-2 with its photon counting technology has shown superior accuracy in comparison to GEDI over bare soils and short vegetation (< 5m), while GEDI was more accurate for tall vegetation across all height ranges (Liu et al., 2021). However, the assessment in the study of Liu et al., 2021 was conducted over geolocation-corrected GEDI and ICESat-2 measurements. Without geolocation-correction, ICESat-2 which shows good agreement with aireborne LiDAR estimates for medium heights (Queinnec et al., 2021), should also be preferred to GEDI acquisitions for this height range due its lower geolocation errors (<5m; Neuenschwander et al., 2022) in comparison to GEDI (geolocation error of 10 m; Dubayah, Ralph et al., 2021).

Over tall trees (e.g. > 25m), GEDI has overall better accuracy (Liu et al., 2021). However, recent studies have shown that GEDI's accuracy is affected by several factors, including the laser used, the processing algorithm used to extract the metrics, as well as acquisition time (Fayad et al., 2021a; Lahssini et al., 2022; Liu et al., 2021). Regarding these two factors and given the limited number of available acquisitions over the tropical forests of Ghana, all available GEDI acquisitions were used in this study, however, over larger areas, a better approach would be to use only high-powered shots. Finally, while both of our proposed models (namely α-2MDU and Hy-TeC) have shown better agreement to ALS than current state-of-the-art, the fact remains that neither GEDI nor models calibrated with GEDI data are capable of retrieving the forest's height dynamics ranges. This finding is also in line with several studies that evaluated the capacity of GEDI to retrieve the canopy heights in dense forests with very heterogeneous compositions (different ages, species, etc.) (Lahssini et al., 2022; Morin et al., 2022).

*5.3. Model Architectures*

Regarding model architecture, our proposed approach differs from current state-of-the-art methods (Lang et al., 2022; Morin et al., 2022; Potapov et al., 2021) in that our models are trained on larger images and they are tasked with creating a continuous height map of the same area, regardless of the density and localization of the GEDI acquisitions within. In contrast, previous approaches used as model input either vegetation indices at a given GEDI's location (Li et al., 2020; Potapov et al., 2021), or small windows that cover GEDI's location and neighboring pixels (Lang et al., 2022). Processing larger images as input during training provides several benefits that can lead to improved performance. First, by leveraging the spatial information within the image, the models can better understand the overall structure of the image and generate more accurate reconstructions (Long et al., 2015). Additionally, processing larger images allows the models to learn more complex and abstract features, which can further improve the reconstruction quality. Second, training on larger images can help prevent overfitting and improve generalization performance by exposing the models to more variations and noise in the training data (Pathak et al., 2016).

Regarding the performance of α-2MDU and Hy-TeC, the results in this study showed that both models were performant in estimating the heights, over the tall tropical forests in central and the south of Ghana, as well as over Savannas. Nonetheless, Hy-TeC, overall, outperformed α-2MDU. Hy-TeC has on average better ground sampling distance, less noise and has lower RMSE when compared to GEDI, ALS and VHR-HM. The difference between the two models is attributed to two main factors. First, Hy-TeC uses a transformer encoder while α-2MDU uses a ConvNet based encoder. This difference is significant, as in order to obtain long-range dependencies between pixels in ConvNet-based networks, it is required to downscale the image or apply a large receptive field, leading to some loss of the finer details in the image and reducing the model's ability to capture complex structures. In contrast, transformer models such as ViTs can capture global information without downsampling or losing details, making them better suited for tasks requiring high resolution and fine-grained features (Dosovitskiy et al., 2020). The second key difference between the two models lies in the use of S1 data. Indeed, the study of Schwartz et al., 2022 showed that a model using both S1 and S2 data is significantly more accurate than a model using either dataset. Nonetheless, speckle noise as well as terrain geometry can introduce artifacts in the estimated height maps from S1 data. The effect of these can be observed in the height estimates of α-2MDU, which in several instances, overestimated the heights in sparse vegetated regions. Hy-TeC, on the other hand, does not have these limitations as it only relies on S2 data as its main input. Hy-TeC, however, has two main limitations. First, it requires height estimates provided from two pre-trained models at lower resolutions to effectively learn an inductive bias in a timely manner. Second, Hy-TeC, as all transformer-based models, require significantly more space than their ConvNets counterparts, which might limit their adoption.

*5.4. Loss Functions*

One of the key contributions of this study is the use of two loss functions to optimize our models. The first loss, a cross entropy loss, is obtained by discretizing the output-space into n-overlapping bins, while the second loss, a regression loss, is used directly on the regression estimates. This approach allowed the models to significantly reduce the underestimation effect that is usually observed when estimating the height of very tall trees. In fact, our two models that employ such loss, showed in general higher tree height estimates over areas with very tall trees in comparison to the estimates obtained in the studies of Potapov et al., 2021 and Lang et al., 2022. Classifying the height range before estimating the height value has two main advantages. First, by constraining the range of heights that can be estimated for a given pixel, the model is allowed more latitude to learn the differences between shorter and taller canopies. Moreover, classification loss can also be weighted, which acts as a counterbalance for underrepresented height classes. By assigning a higher weight to the minority classes, the classification loss will place more emphasis on correctly predicting these classes, effectively compensating for their underrepresentation in the training data. Indeed, in our training data, only a small fraction of the available GEDI footprints were acquired over tall trees (less than 1%), yet model accuracy on the height estimation of these GEDI footprints was similar to those with significantly higher count.

Finally, in this study we replaced the de-facto standard of regression losses, that is the L1/L2 losses, by an adaptive loss that constrains the value of the residuals to an automatically set bound. This loss which reduces the effects of outliers was used as a simple alternative to a more elaborate filtering scheme of GEDI acquisitions and resulted in better quality maps in comparison to the L1/L2 losses. Nonetheless, while the adaptive loss function's ability to reduce the effect of outliers is undeniable, we are uncertain of its effects on the inliers that represent hard cases, such as the case of very tall trees present within a forest of significantly shorter ones. As such, more research needs to be conducted to quantify the advantages of using such loss in comparison to the unbounded losses such as L1/L2 coupled with a better filtered GEDI training dataset.

## Appendix A. Implementation of the UNET encoder block ($CEB_i$):

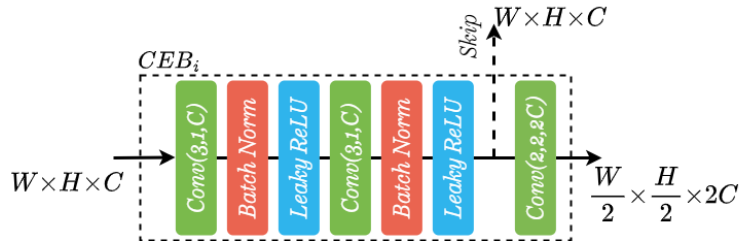

Figure A.20. A U-Net encoder block ($CEB_i$). *Conv(K, S, F)* represents a convolution operation with a kernel of size K, stride S and F filters (filters and channels are mentioned interchangeably). *Batch Norm* is a batch normalization layer. *Leaky ReLU* is a leaky rectified linear unit.

## Appendix B. Implementation of the UNET decoder block ($CDB_i$):

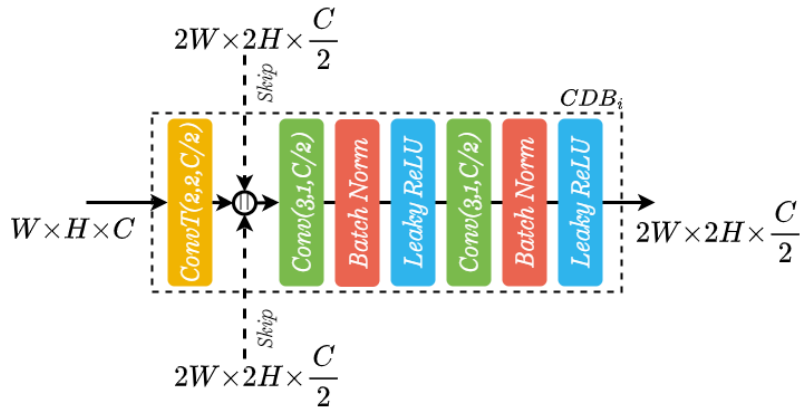

Figure B.21. Implemented decoder block ($CDB_i$). *ConvT(K, S, F)* represents a transpose convolution operation with a kernel of size K, stride S, and F filters. The '||' operation represents a channel-wise concatenation.

**Appendix C. Barron, 2017 Adaptive Loss**

The adaptive loss is a two-parameter loss function that generalizes many one-parameter loss functions (L1 or MAE, L2 or MSE, Cauchy/Lorentzian, Geman-McClure, Welsch/Leclerc, generalized Charbonnier, etc.), and is defined as follows:

$$\ell_{ar}(Y, Y') = \frac{|\alpha - 2|}{\alpha}\left(\left(\frac{\left(\frac{(Y-Y')}{c}\right)^2}{|\alpha - 2|} + 1\right)^{\frac{\alpha}{2}} - 1\right) \quad \text{C.11}$$

Where $\ell_{ar}$ is the robust-adaptive loss function, $Y$ is the vector of target values to estimate and $Y'$ the vector of truth values, $\alpha$ is a shape parameter that controls the robustness of the loss and c is a scaling parameter. Both parameters are learned during training.

**Appendix D. Image pre-processing for ViT based architectures**

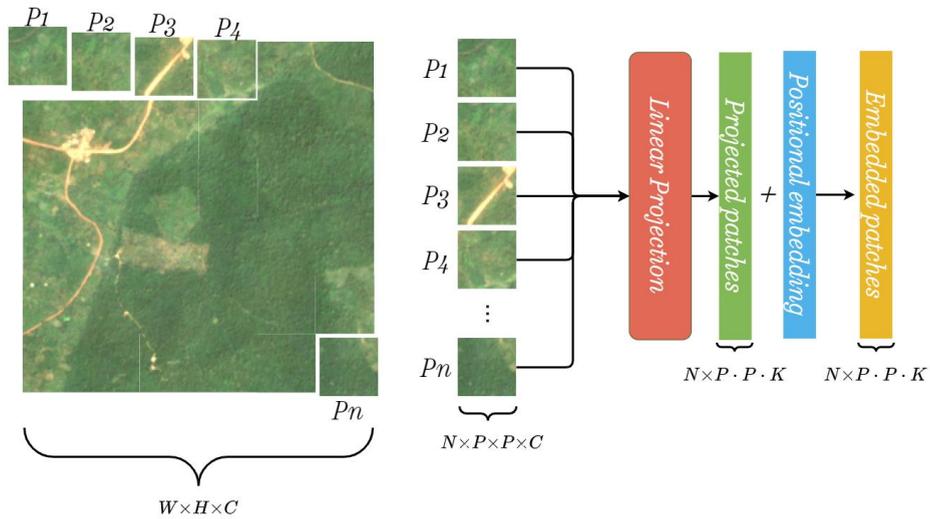

Figure D.22. Visualizing the patching, projection, and positional embedding steps in the image pre-processing pipeline for ViT based models.

**Appendix E. The implementation of Hy-TeC reprojection blocks ($RB_i$)**

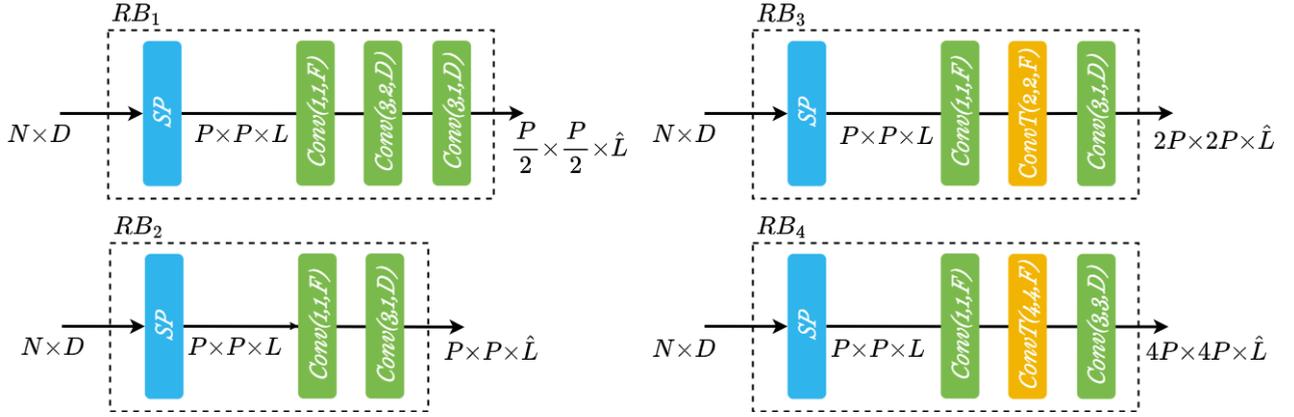

Figure E.23. The implementation of the four reprojection blocks ($RB_i$).

## Appendix F. The implementation of Hy-TeC decoder blocks ($DB_i$)

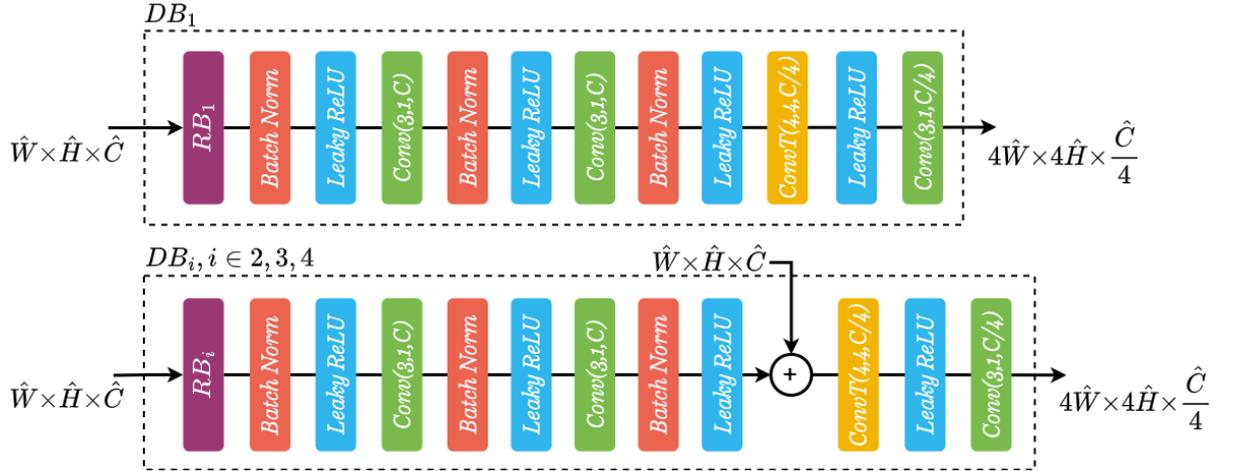

Figure F.24. The implementation of the four decoder blocks ($DB_i$).

## Appendix G. Selected grid cells used for the training/validation of the proposed models

| Set | $RH_{98}$ range | Ratio of $RH_{98}$ within a cell | Cell Count | Number of duplications |
|---|---|---|---|---|
| 1 | $RH_{98} \leq 5$m | $> 50\%$ | 1152 | 0 |
| 2 | $5\text{ m} < RH_{98} \leq 10\text{m}$ | $> 25\%$ | 327 | 1 |
| 3 | $10\text{ m} < RH_{98} \leq 15\text{m}$ | $> 10\%$ | 469 | 0 |
| 4 | $15\text{ m} < RH_{98} \leq 20\text{m}$ | $> 10\%$ | 149 | 3 |
| 5 | $20\text{ m} < RH_{98} \leq 25\text{m}$ | $> 5\%$ | 133 | 4 |
| 6 | $25\text{ m} < RH_{98} \leq 30\text{m}$ | $> 2.5\%$ | 139 | 4 |
| 7 | $30\text{ m} < RH_{98} \leq 35\text{m}$ | $> 2.5\%$ | 52 | 8 |
| 8 | $35\text{ m} < RH_{98} \leq 40\text{m}$ | $> 2.5\%$ | 47 | 8 |
| 9 | $RH_{98} \geq 40$ | $> 2.5\%$ | 91 | 4 |

Table G.3. The number of 768×768 grid cells based on the ratio of $RH_{98}$/cell and the $RH_{98}$ height ranges.

# Appendix H. Summary statistical metric definitions

$$r = \frac{\sum_{i=1}^{n}(y_i - \hat{y})(\hat{y} - \bar{\hat{y}})}{\sqrt{\sum_{i=1}^{n}(y_i - \bar{y})^2} \sqrt{\sum_{i=1}^{n}(\hat{y}_i - \bar{\hat{y}})}} \qquad 12$$

$$RMSE = \sqrt{\frac{1}{n} \cdot \sum_{i=1}^{n}(y_i - \hat{y}_i)^2} \qquad 13$$

$$RMSPE = 100 \cdot \sqrt{\frac{1}{n} \cdot \sum_{i=1}^{n}\left(\frac{y_i - \hat{y}_i}{y_i}\right)^2} \qquad 14$$

$$SDSD = (SD_s - SD_m)^2 \qquad 15$$

$$LCS = 2SD_s SD_m (1 - r) \qquad 16$$

Where $y_i$ is the target value (i.e. $RH_{98}$), $\hat{y}_1$ the estimated value, $\bar{y}_1$ is the mean of all observed values, $\bar{\hat{y}}$ is the mean of all estimated values, n is the sample size and $SD_s$ and $SD_m$ are respectively the Standard deviation of the measured heights by GEDI and the standard deviation of the estimated heights by the models. They are defined as follows:

$$SD_m = \sqrt{\frac{\sum_{i=1}^{n}(y_i - \bar{y})}{N}} \qquad 17$$

$$SD_s = \sqrt{\frac{\sum_{i=1}^{n}(\hat{y}_i - \bar{\hat{y}})}{N}} \qquad 18$$

# Appendix I. Degradation effect observed from 2MDU's generated height maps

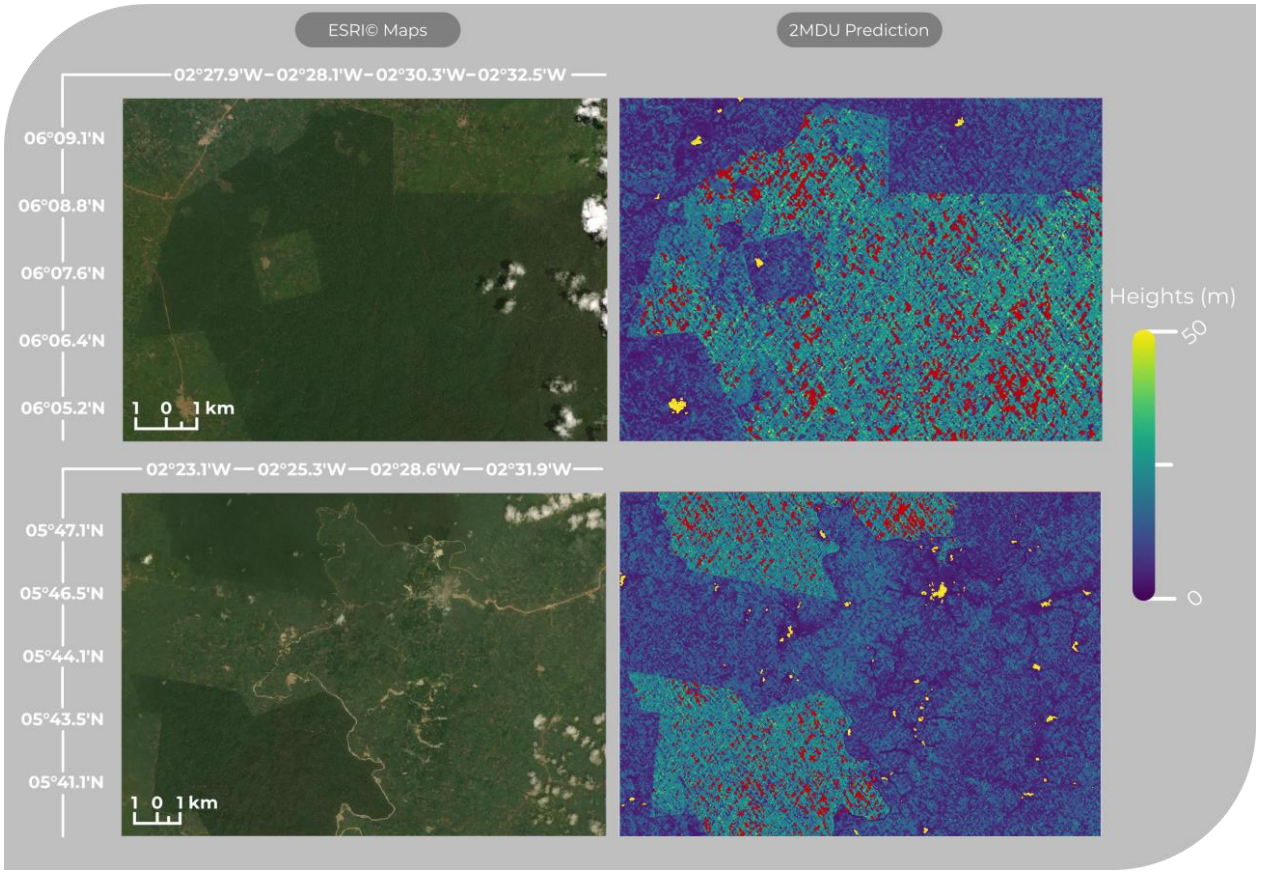

Figure I.25. Illustration of the degraded performance (red polygons) of the 2MDU model over two forest patches.

## Appendix J. Comparison between the residual and gradients of the adaptive loss and Huber loss

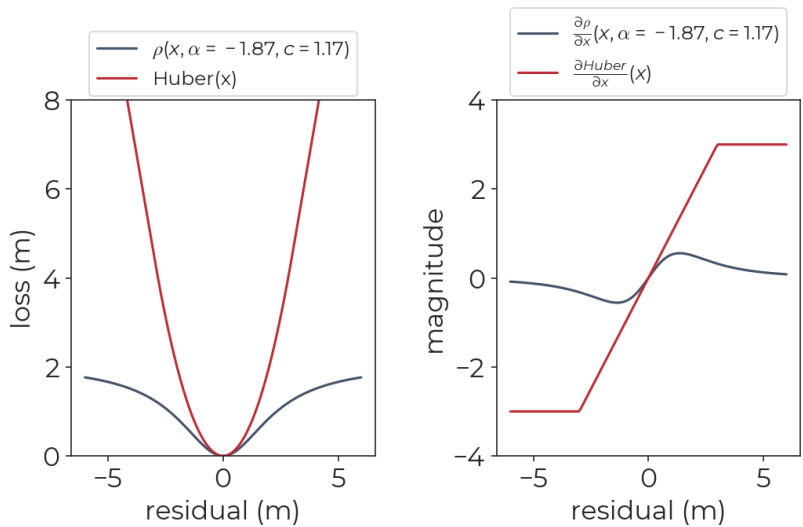

Figure J.26. The adaptive (ρ) and Huber loss functions (left) and their gradients (right).

**Appendix K. Distribution of GSI from the four compared height products**

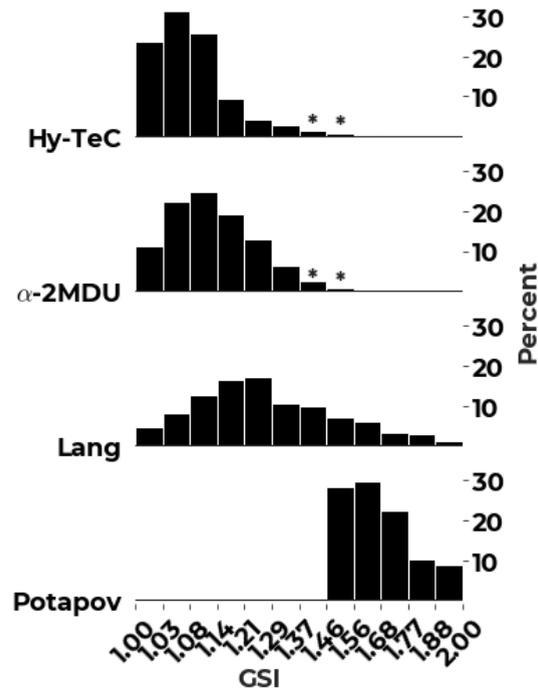

Figure K.27. GSI distribution of the 5000 256×256 height patches from the four compared height models over Ghana. '*' represents less than 1% of patches.